\documentclass[journal,twoside,web]{ieeecolor}

\usepackage{generic}
\usepackage{cite}
\usepackage{amsmath,amssymb,amsfonts}
\usepackage{algorithmic}
\usepackage{graphicx}
\usepackage{textcomp}

\usepackage{bm}
\usepackage[per-mode=symbol]{siunitx}
\usepackage{tikz}
\usetikzlibrary{matrix, calc}
\usepackage{varwidth}
\usepackage{nicefrac}
\usepackage{xurl}
\usepackage{subcaption}
\usepackage{mathtools}
\usepackage{multirow}
\usepackage[hidelinks]{hyperref}
\usepackage{comment}  
\usepackage{fancyhdr}

\let\labelindent\relax
\usepackage[inline]{enumitem}

\newcommand{\euler}{\mathrm{e}}

\DeclareMathOperator*{\argmin}{arg\,min}
\DeclareMathOperator*{\diag}{diag}
\DeclareMathOperator*{\blockdiag}{blockdiag}
\DeclareMathOperator*{\st}{s.t.}
\DeclareMathOperator*{\MSE}{MSE}
\DeclareMathOperator*{\MNAE}{MNAE}

\usepackage{xspace}
\newcommand*{\eg}{e.g.\@\xspace}
\newcommand*{\ie}{i.e.\@\xspace}

\newcommand*{\wrt}{w.r.t.\@\xspace}
\newcommand*{\resp}{resp.\@\xspace}
\newcommand*{\ST}{s.t.\@\xspace}
\newcommand*{\ms}{M.Sc.\@\xspace}
\newcommand*{\phd}{Ph.D.\@\xspace}
\newcommand*{\dr}{Dr.\@\xspace}
\newcommand*{\prof}{Prof.\@\xspace}

\usepackage[acronym,toc,symbols,nogroupskip,sort=none]{glossaries-extra}

\glsxtrnewsymbol[description={$\in \mathbb N$ number of degrees of freedom}]{n-dof}{\ensuremath{n}}
\glsxtrnewsymbol[description={$\in \mathbb R^{\gls{n-dof}}$ joint positions, velocities, accelerations}]{j-positions}{\ensuremath{\bm q, \dot{\bm q}, \ddot{\bm q}}}
\glsxtrnewsymbol[description={$\in \mathbb R^{\gls{n-dof} \times \gls{n-dof}}$ inertia matrix}]{inertia}{\ensuremath{\bm H}}
\glsxtrnewsymbol[description={$\in \mathbb R^{1 \times \gls{n-dof}}$ $j$-th row of $\gls{inertia}$}]{inertia-j}{\ensuremath{\bm h_j}}
\glsxtrnewsymbol[description={$\in \mathbb R^{\gls{n-dof} \times \gls{n-dof}}$ centrifugal and Coriolis matrix}]{coriolis}{\ensuremath{\bm C}}
\glsxtrnewsymbol[description={$\in \mathbb R^{1 \times \gls{n-dof}}$ $j$-th row of $\gls{coriolis}$}]{coriolis-j}{\ensuremath{\bm c_j}}
\glsxtrnewsymbol[description={$\in \mathbb R^{\gls{n-dof}}$ friction torques}]{friction}{\ensuremath{\bm f}}
\glsxtrnewsymbol[description={$\in \mathbb R^{\gls{n-dof}}$ gravity torques}]{gravity}{\ensuremath{\bm g}}
\glsxtrnewsymbol[description={$\in \mathbb R^{\gls{n-dof}}$ joint torques, currents}]{torques}{\ensuremath{\bm\tau, \bm v}}
\glsxtrnewsymbol[description={$\in \mathbb R^{\gls{n-dof} \times \gls{n-dof}}$ diagonal matrix of joint motor drive gains}]{mdg}{\ensuremath{\bm K}}
\glsxtrnewsymbol[description={quantity $(\bullet)$ for the $j$-th joint}]{jth}{\ensuremath{\bullet_j}}

\glsxtrnewsymbol[description={$\in \mathbb R_+$ mass}]{mass}{\ensuremath{m}}
\glsxtrnewsymbol[description={$\in \mathbb R^3$ center of mass}]{com}{\ensuremath{\bm r}}
\glsxtrnewsymbol[description={$\in \mathbb R^{3 \times 3}$ inertia tensor relative to the principal axes of inertia}]{inertia-tensor-pai}{\ensuremath{\check{\bm I}}}
\glsxtrnewsymbol[description={$\in \mathbb R^{3 \times 3}$ inertia tensor relative to the kinematic frame}]{inertia-tensor}{\ensuremath{\bm I}}
\glsxtrnewsymbol[description={$\in \mathbb R^{10\gls{n-dof}}$ dynamic parameters}]{d-parameters}{\ensuremath{\bm\pi}}
\glsxtrnewsymbol[description={quantity \ensuremath{(\bullet)} for the $i$-th link}]{ith}{\ensuremath{\bullet_i}}
\glsxtrnewsymbol[description={$\in \mathbb R^{10(\gls{n-dof}-1)}$ dynamic parameters of links $1, \ldots, \gls{n-dof}$}]{d-parameters-1-to-nm1}{\ensuremath{\bm\pi_{1:n-1}}}

\glsxtrnewsymbol[description={$\in \mathbb N$ number of dynamic coefficients}]{n-d-coefficients}{\ensuremath{p}}
\glsxtrnewsymbol[description={$\in \mathbb R^{\gls{n-d-coefficients}}$ dynamic coefficients}]{d-coefficients}{\ensuremath{\bm\xi}}
\glsxtrnewsymbol[description={$\in \mathbb R^{\gls{n-d-coefficients}\gls{n-dof}}$ current-level dynamic coefficients}]{c-d-coefficients}{\ensuremath{\bm\chi}}
\glsxtrnewsymbol[description={$\in \mathbb R^{\gls{n-dof} \times 10\gls{n-dof}}$ regressor}]{regressor}{\ensuremath{\bm P}}
\glsxtrnewsymbol[description={$\in \mathbb R^{\gls{n-dof} \times 10(\gls{n-dof}-1)}$ columns of \gls{regressor} related to $\gls{d-parameters}_1, \ldots, \gls{d-parameters}_{n-1}$}]{regressor-1-to-nm1}{\ensuremath{\bm P_{1:n-1}}}
\glsxtrnewsymbol[description={$\in \mathbb R^{\gls{n-dof} \times 10}$ columns of \gls{regressor} related to $\gls{d-parameters}_n$}]{regressor-n}{\ensuremath{\bm P_n}}
\glsxtrnewsymbol[description={$\in \mathbb R^{\gls{n-dof} \times \gls{n-d-coefficients}}$ minimal regressor}]{minimal-regressor}{\ensuremath{\bm Y}}
\glsxtrnewsymbol[description={$\in \mathbb R^{1 \times \gls{n-d-coefficients}}$ $j$-th row of $\gls{minimal-regressor}$}]{minimal-regressor-j}{\ensuremath{\bm y_j}}
\glsxtrnewsymbol[description={$\in \mathbb R^{\gls{n-dof} \times \gls{n-d-coefficients}\gls{n-dof}}$ current-level minimal regressor}]{c-minimal-regressor}{\ensuremath{\bm U}}
\glsxtrnewsymbol[description={$\in \mathbb R^{1 \times \gls{n-d-coefficients}\gls{n-dof}}$ $j$-th row of $\gls{c-minimal-regressor}$}]{c-minimal-regressor-j}{\ensuremath{\bm u_j}}
\glsxtrnewsymbol[description={diagonal matrix of weights for WLSE}]{weights}{\ensuremath{\bm W}}
\glsxtrnewsymbol[description={$\in \mathbb R^{\gls{n-dof}}$ variance of the $k$-th sample in $\underline{\bm\tau}$ or $\underline{\bm v}$}]{variance}{\ensuremath{\bm\sigma_k^2}}

\glsxtrnewsymbol[description={$\in \mathbb R^{\gls{n-dof}}$ joint currents due to friction}]{c-f}{\ensuremath{\bm v_f}}
\glsxtrnewsymbol[description={$\in \mathbb R^{\gls{n-dof}}$ Coulomb friction offset}]{f-o}{\ensuremath{\bm f_o}}
\glsxtrnewsymbol[description={$\in \mathbb R^{\gls{n-dof}}$ viscous friction}]{f-v}{\ensuremath{\bm f_v}}
\glsxtrnewsymbol[description={$\in \mathbb R^{\gls{n-dof}}$ Coulomb friction coefficient}]{f-c}{\ensuremath{\bm f_c}}
\glsxtrnewsymbol[description={$\in \mathbb R^{\gls{n-dof}}$ nonlinear friction multiplicative factor}]{f-mult}{\ensuremath{\bm\delta}}
\glsxtrnewsymbol[description={$\in \mathbb R^{\gls{n-dof}}$ nonlinear friction additive factor}]{f-add}{\ensuremath{\bm\nu}}
\glsxtrnewsymbol[description={set of friction parameters}]{f-set}{\ensuremath{\bm\psi}}
\glsxtrnewsymbol[description={$\in \mathbb R^{\gls{n-dof}}$ nonlinear friction torques \gls{friction} evaluated on $\gls{f-set}$}]{torques-f-set}{\ensuremath{\gls{friction}_{\gls{f-set}}}}
\glsxtrnewsymbol[description={$\in \mathbb R^{\gls{n-dof}}$ nonlinear friction currents evaluated on $\gls{f-set}$}]{c-f-set}{\ensuremath{\bm v_{\bm\psi}}}
\glsxtrnewsymbol[description={$\in \mathbb R^{(\gls{n-d-coefficients}-3)\gls{n-dof}}$ $\gls{c-d-coefficients}$ excluding \gls{f-o}, \gls{f-v}, \gls{f-c}}]{c-d-coefficients-f}{\ensuremath{\bm\chi_f}}
\glsxtrnewsymbol[description={$\in \mathbb R^{\gls{n-dof} \times (\gls{n-d-coefficients}-3)\gls{n-dof}}$ \gls{c-minimal-regressor} excluding columns related to \gls{f-o}, \gls{f-v}, \gls{f-c}}]{c-minimal-regressor-f}{\ensuremath{\bm U_f}}
\glsxtrnewsymbol[description={$\in \mathbb R^{1 \times (\gls{n-d-coefficients}-3)\gls{n-dof}}$ $j$-th row of \gls{c-minimal-regressor-f}}]{c-minimal-regressor-f-j}{\ensuremath{\bm u_{f,j}}}

\glsxtrnewsymbol[description={$\in \mathbb N$ number of unknown payload dynamic parameters}]{n-u-p-d-parameters}{\ensuremath{c}}
\glsxtrnewsymbol[description={quantity $(\bullet)$ for payload \wrt payload origin frame}]{p-origin-frame}{\ensuremath{\bullet_l}}
\glsxtrnewsymbol[description={quantity $(\bullet)$ for payload \wrt \gls{n-dof}-th link frame}]{p-framev}{\ensuremath{\bullet_L}}
\glsxtrnewsymbol[description={$\in SO(3)$ rotation of frame $l$ \wrt \gls{n-dof}-th link frame}]{p-rotation}{\ensuremath{\bm R^l_n}}
\glsxtrnewsymbol[description={$\in \mathbb R^3$ translation of frame $l$ \wrt \gls{n-dof}-th link frame}]{p-translation}{\ensuremath{\bm t^l_n}}
\glsxtrnewsymbol[description={$\in SE(3)$ transformation of frame $l$ \wrt \gls{n-dof}-th link frame}]{p-transformation}{\ensuremath{\bm T^l_n}}
\glsxtrnewsymbol[description={$\in \mathbb R^{\gls{n-u-p-d-parameters}}$ unknown payload dynamic parameters}]{u-p-d-parameters}{\ensuremath{\bm\pi_{uL}}}
\glsxtrnewsymbol[description={$\in \mathbb R^{10-\gls{n-u-p-d-parameters}}$ known payload dynamic parameters}]{k-p-d-parameters}{\ensuremath{\bm\pi_{kL}}}
\glsxtrnewsymbol[description={$\in \mathbb R^{\gls{n-dof} \times \gls{n-u-p-d-parameters}}$ regressor \wrt \gls{u-p-d-parameters}}]{regressor-u-p}{\ensuremath{\gls{regressor}_{uL}}}
\glsxtrnewsymbol[description={$\in \mathbb R^{\gls{n-dof} \times (10-\gls{n-u-p-d-parameters})}$ regressor \wrt \gls{k-p-d-parameters}}]{regressor-k-p}{\ensuremath{\gls{regressor}_{kL}}}
\glsxtrnewsymbol[description={$\in \mathbb R^{1 \times \gls{n-u-p-d-parameters}}$ $j$-th row of \gls{regressor-u-p}}]{regressor-u-p-j}{\ensuremath{\bm p_{uL,j}}}
\glsxtrnewsymbol[description={$\in \mathbb R^{1 \times (10-\gls{n-u-p-d-parameters})}$ $j$-th row of \gls{regressor-k-p}}]{regressor-k-p-j}{\ensuremath{\bm p_{kL,j}}}
\glsxtrnewsymbol[description={$\in \mathbb R^{\gls{n-dof}}$ joint torques due to arm motion only}]{torques-arm}{\ensuremath{\bm\tau_{arm}}}
\glsxtrnewsymbol[description={$\in \mathbb R^{\gls{n-dof}}$ joint torques due to payload}]{torques-p}{\ensuremath{\bm\tau_L}}

\glsxtrnewsymbol[description={$\in \mathbb N$ number of Fourier components in exciting trajectories}]{n-fourier-components}{\ensuremath{F}}
\glsxtrnewsymbol[description={$\in \mathbb N$ number of samples acquired with exciting trajectories}]{n-samples}{\ensuremath{M}}
\glsxtrnewsymbol[description={$\in \mathbb N$ number of samples acquired in static conditions}]{n-static-samples}{\ensuremath{N}}
\glsxtrnewsymbol[description={$\in \mathbb R^{\gls{n-dof}}$ excitation trajectory parameters}]{et-center}{\ensuremath{\bm q_0, \bm a_k, \bm b_k}}

\glsxtrnewsymbol[description={quantity $(\bullet)$ for trajectories executed without payload}]{w-p}{\ensuremath{\bullet^a}}
\glsxtrnewsymbol[description={quantity $(\bullet)$ for trajectories executed with payload}]{wo-p}{\ensuremath{\bullet^b}}
\newcommand{\regressormdg}{\ensuremath{\bm S}}
\glsxtrnewsymbol[description={$\in \mathbb R^{M \times (\gls{n-d-coefficients}+\gls{n-u-p-d-parameters}-2)}$ regressor to estimate $K_j$}]{regressor-mdg-j}{\ensuremath{{\regressormdg}_j}}
\glsxtrnewsymbol[description={$\in \mathbb R^{\gls{n-d-coefficients}+\gls{n-u-p-d-parameters}-2}$ vector including arm coefficients and $K_j$}]{vector-mdg-j}{\ensuremath{\bm\zeta_j}}
\glsxtrnewsymbol[description={$\in \mathbb R^{1 \times (\gls{n-d-coefficients}+\gls{n-u-p-d-parameters}-2)}$ decision vector to extract $K_j$ from $\gls{vector-mdg-j}$}]{decision-vector-j}{\ensuremath{\bm\mu_j}}
\glsxtrnewsymbol[description={$\in \mathbb N$ number of identifiable parameters in \gls{vector-mdg-j}}]{n-identifiable-parameters-j}{\ensuremath{d_j}}
\newcommand{\identifiableparameters}{\ensuremath{\bm\phi}}
\glsxtrnewsymbol[description={$\in \mathbb R^{\gls{n-identifiable-parameters-j}}$ identifiable parameters in \gls{vector-mdg-j}}]{identifiable-parameters-j}{\ensuremath{\identifiableparameters_j}}
\newcommand{\unidentifiableparameters}{\ensuremath{\bm\varphi}}
\glsxtrnewsymbol[description={$\in \mathbb R^{\gls{n-d-coefficients}+\gls{n-u-p-d-parameters}-2-\gls{n-identifiable-parameters-j}}$ unidentifiable parameters in \gls{vector-mdg-j}}]{unidentifiable-parameters-j}{\ensuremath{\unidentifiableparameters_j}}
\glsxtrnewsymbol[description={$\in \mathbb R^{\gls{n-samples} \times \gls{n-identifiable-parameters-j}}$ linearly independent columns of \gls{regressor-mdg-j}}]{regressor-mdg-j-li}{\ensuremath{\bm S_{j,\phi}}}
\glsxtrnewsymbol[description={$\in \mathbb R^{\gls{n-samples} \times (\gls{n-d-coefficients}+\gls{n-u-p-d-parameters}-2-\gls{n-identifiable-parameters-j})}$ linearly dependent columns of \gls{regressor-mdg-j}}]{regressor-mdg-j-ld}{\ensuremath{\bm S_{j,\varphi}}}
\glsxtrnewsymbol[description={$\in \mathbb R^{\gls{n-samples} \times \gls{n-samples}}$ orthogonal matrix of \gls{regressor-mdg-j-li}, \gls{regressor-mdg-j-ld}}]{q-mdg-j-li}{\ensuremath{\bm{\mathcal Q}_{j,\phi}, \bm{\mathcal Q}_{j,\varphi}}}
\newcommand{\rmdg}{\ensuremath{\bm{\mathcal R}}}
\glsxtrnewsymbol[description={$\in \mathbb R^{\gls{n-samples} \times \gls{n-identifiable-parameters-j}}$ upper triangular matrix of \gls{regressor-mdg-j-li}}]{r-mdg-j-li}{\ensuremath{\rmdg_{j,\phi}}}
\glsxtrnewsymbol[description={$\in \mathbb R^{\gls{n-samples} \times (\gls{n-d-coefficients}+\gls{n-u-p-d-parameters}-2-\gls{n-identifiable-parameters-j})}$ upper triangular matrix of \gls{regressor-mdg-j-ld}}]{r-mdg-j-ld}{\ensuremath{\rmdg_{j,\varphi}}}
\glsxtrnewsymbol[description={$\in \mathbb R^{\gls{n-identifiable-parameters-j}}$ regrouping vector of \gls{vector-mdg-j}}]{regrouping-vector-j}{\ensuremath{\bm\beta_j}}

\glsxtrnewsymbol[description={$\in\mathbb R^{3 \times 3}$ skew-symmetric operator applied to vector \ensuremath{(\bullet)}}]{sso}{\ensuremath{[\bullet]_\times}}
\glsxtrnewsymbol[description={estimation of quantity \ensuremath{(\bullet)}}]{estimation}{\ensuremath{\hat{\bullet}}}
\glsxtrnewsymbol[description={$\in \mathbb R^{Mx \times y}$ column-stack of quantity \ensuremath{(\bullet) \in \mathbb R^{x \times y}}}]{stack}{\ensuremath{\underline{\bullet}}}
\glsxtrnewsymbol[description={lower/upper bounds on quantity $(\bullet)$}]{lb}{\ensuremath{\bullet^-, \bullet^+}}
\glsxtrnewsymbol[description={$\in \mathbb R_+$ improvement factor of $\MSE(\bm v_j,\hat{\bm v}_j)$ \wrt \cite{gaz_model-based_2018}}]{improvement-factor}{\ensuremath{\eta}}

\renewcommand{\baselinestretch}{0.97}

\definecolor{lgblue}{RGB}{0, 113.985, 188.955}
\definecolor{lgred}{RGB}{216.750, 82.875, 24.990}
\definecolor{lgyellow}{RGB}{236.895, 176.970, 31.875}

\newcommand{\defineleg}[2]{
    \newcommand{#1}{\raisebox{2pt}{\tikz{\draw[#2, solid, line width=0.9pt](0,0)--(6mm,0);}}}
}
\defineleg{\legblue}{lgblue}
\defineleg{\legred}{lgred}
\defineleg{\legyellow}{lgyellow}
\newcommand{\boxgray}{\raisebox{2pt}{\tikz[baseline=(current bounding box.center)]{\fill[black!7.5] (0,0) rectangle (6mm,2mm);}}}

\makeatletter
\def\endthebibliography{%
  \def\@noitemerr{\@latex@warning{Empty `thebibliography' environment}}%
  \endlist
}
\makeatother

\definecolor{bostonuniversityred}{rgb}{0.8, 0.0, 0.0}
\definecolor{ao(english)}{rgb}{0.0, 0.5, 0.0}
\definecolor{blue(pigment)}{rgb}{0.2, 0.2, 0.6}
\tikzset{
redcross/.style={
    path picture={\draw[bostonuniversityred] (path picture bounding box.south east) -- (path picture bounding box.north west) (path picture bounding box.south west) -- (path picture bounding box.north east);}
},
greencross/.style={
    path picture={\draw[ao(english)] (path picture bounding box.south east) -- (path picture bounding box.north west) (path picture bounding box.south west) -- (path picture bounding box.north east);}
},
}

\def\BibTeX{{\rm B\kern-.05em{\sc i\kern-.025em b}\kern-.08em
    T\kern-.1667em\lower.7ex\hbox{E}\kern-.125emX}}
\begin{document}
\title{The Dynamic Model of the UR10 Robot and its ROS2 Integration}
\author{
Vincenzo Petrone,
Enrico Ferrentino, and
Pasquale Chiacchio
\thanks{
Authors are with the Department of Information Engineering, Electrical Engineering and Applied Mathematics (DIEM), University of Salerno, 84048 Fisciano, Italy (e-mail: vipetrone@unisa.it, eferrentino@unisa.it, pchiacchio@unisa.it).
}
\thanks{
\textcopyright 2025 IEEE.
Personal use of this material is permitted.
Permission from IEEE must be obtained for all other uses, in any current or future media, including reprinting/republishing this material for advertising or promotional purposes, creating new collective works, for resale or redistribution to servers or lists, or reuse of any copyrighted component of this work in other works.
}
}

\maketitle

\begin{abstract}
This paper presents the full dynamic model of the UR10 industrial robot.
A triple-stage identification approach is adopted to estimate the manipulator's dynamic coefficients.
First, linear parameters are computed using a standard linear regression algorithm.
Subsequently, nonlinear friction parameters are estimated according to a sigmoidal model.
Lastly, motor drive gains are devised to map estimated joint currents to torques.
The overall identified model can be used for both control and planning purposes, as the accompanied ROS2 software can be easily reconfigured to account for a generic payload.
The estimated robot model is experimentally validated against a set of exciting trajectories and compared to the state-of-the-art model for the same manipulator, achieving higher current prediction accuracy (up to a factor of 4.43) and more precise motor gains.
The related software is available at \url{https://codeocean.com/capsule/8515919/tree/v2}.
\end{abstract}


\section{Introduction}

\IEEEPARstart{M}{odeling} manipulator dynamics is critical for a broad spectrum of robotic applications, ranging from high-performance model-based control \cite{lu_inverse_2024} to trajectory planning \cite{petrone_time-optimal_2022} and human-robot interaction \cite{gaz_model-based_2018, zhang_sensorless_2019, xing_dynamic_2024}.

In industrial settings, the effectiveness of planning and control algorithms heavily relies on the accuracy of the underlying dynamic model \cite{yang_highly_2023}.
Therefore, obtaining a reliable estimation of manipulator dynamics is essential in all the aforementioned applications \cite{swevers_dynamic_2007}.

\subsection{Background} \label{sec:background}

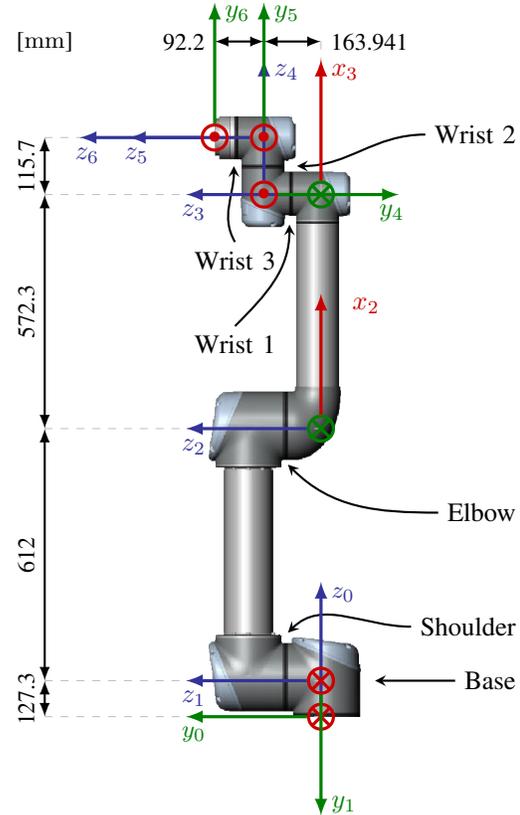
\begin{figure}
\centering

\begin{tikzpicture}
\newcommand{\labelpos}{0.9}
\newcommand{\framearrowlen}{0.225}
\begin{scope}
    \node[anchor=south west, inner sep=0] (image) at (0,0){\includegraphics[width=0.9\columnwidth]{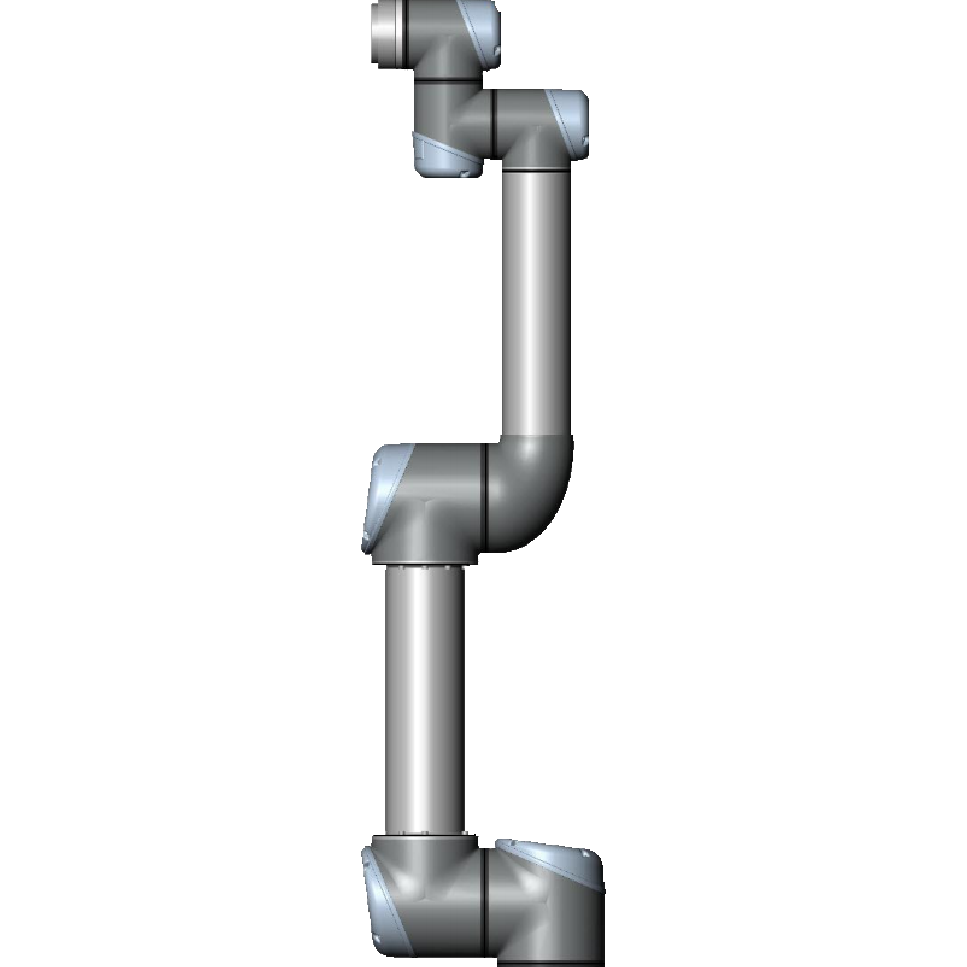}};
    \begin{scope}[x={(image.south east)},y={(image.north west)}]
        \node [anchor=east] (base) at (\labelpos,0.06) {Base};
        \node [anchor=east] (shoulder) at (\labelpos,0.15) {Shoulder};
        \node [anchor=east] (elbow) at (\labelpos,0.34) {Elbow};
        \node [anchor=east] (wrist1) at (0.5, 0.62) {Wrist 1};
        \node [anchor=east] (wrist2) at (\labelpos,0.97) {Wrist 2};
        \node [anchor=east] (wrist3) at (0.5, 0.76) {Wrist 3};
        
        \draw[-stealth, thick, black] (base.west) to (0.65, 0.06);
        \draw[-stealth, thick, black] (shoulder.west) to[out=180, in=45] (0.505, 0.14);
        \draw[-stealth, thick, black] (elbow.west) to[out=180, in=-45] (0.505, 0.42);
        \draw[-stealth, thick, black] (wrist1.north) to[out=45, in=-90] (0.5075, 0.82);
        \draw[-stealth, thick, black] (wrist2.west) to[out=180, in=0] (0.52, 0.9175);
        \draw[-stealth, thick, black] (wrist3.north) to[out=120, in=-90] (0.415, 0.9175);

        \newcommand{\frameZeroPosX}{0.56}
        \newcommand{\frameOnePosY}{0.06}
        \newcommand{\frameTwoPosY}{0.48}
        \newcommand{\frameThreePosY}{0.87}
        \newcommand{\frameFourPosX}{0.465}
        \newcommand{\frameFivePosY}{0.965}
        \newcommand{\frameSixPosX}{0.383}
        
        \newcommand{\arrowPosX}{+0.1}
        \newcommand{\arrowPosY}{1.125}
        \draw[latex-latex, thick, black] (\arrowPosX, 0.0) --node[midway, above, rotate=90, font=\small]{127.3} (\arrowPosX, \frameOnePosY);
        \draw[latex-latex, thick, black] (\arrowPosX, \frameOnePosY) --node[midway, above, rotate=90, font=\small]{612} (\arrowPosX, \frameTwoPosY);
        \draw[latex-latex, thick, black] (\arrowPosX, \frameTwoPosY) --node[midway, above, rotate=90, font=\small]{572.3} (\arrowPosX, \frameThreePosY);
        \draw[latex-latex, thick, black] (\arrowPosX, \frameThreePosY)  --node[midway, above, rotate=90, font=\small]{115.7} (\arrowPosX, \frameFivePosY);
        \draw[latex-latex, thick, black] (\frameSixPosX, \arrowPosY) --node[pos=0, left, font=\small]{92.2} (\frameFourPosX, \arrowPosY);
        \draw[latex-latex, thick, black] (\frameFourPosX, \arrowPosY) --node[pos=1, right, font=\small]{163.941} (\frameZeroPosX, \arrowPosY);
        \node[font=\small] (dim) at (\arrowPosX, \arrowPosY) {[\si{\milli\meter}]};

        \draw[dashed, ultra thin, black!25] (\arrowPosX, 0) -- (\frameZeroPosX, 0) {};
        \draw[dashed, ultra thin, black!25] (\arrowPosX, \frameOnePosY) -- (\frameZeroPosX, \frameOnePosY) {};
        \draw[dashed, ultra thin, black!25] (\arrowPosX, \frameTwoPosY) -- (\frameZeroPosX, \frameTwoPosY) {};
        \draw[dashed, ultra thin, black!25] (\arrowPosX, \frameThreePosY) -- (\frameFourPosX, \frameThreePosY) {};
        \draw[dashed, ultra thin, black!25] (\arrowPosX, \frameFivePosY) -- (\frameSixPosX, \frameFivePosY) {};
        \draw[dashed, ultra thin, black!25] (\frameZeroPosX, \arrowPosY) -- (\frameZeroPosX, \frameThreePosY) {};

        \node[draw, very thick, circle, redcross, bostonuniversityred] (x0) at (\frameZeroPosX, 0) {};
        \draw[-latex, very thick, ao(english), font=\small] (x0) -- node[near end, anchor=north east]{$y_0$} ++(-\framearrowlen, 0);
        \draw[-latex, very thick, blue(pigment), font=\small] (x0) -- node[near end, anchor=south west]{$z_0$} ++(0, \framearrowlen);
        \node[draw, very thick, circle, redcross, bostonuniversityred] (x1) at (\frameZeroPosX, \frameOnePosY) {};
        \draw[-latex, very thick, ao(english)] (x1) -- node[near end, anchor=north west]{$y_1$} ++(0, -\framearrowlen);
        \draw[-latex, very thick, blue(pigment)] (x1) -- node[near end, anchor=north east]{$z_1$} ++(-\framearrowlen, 0);
        \node[draw, very thick, circle, greencross, ao(english)] (y2) at (\frameZeroPosX, \frameTwoPosY) {};
        \draw[-latex, very thick, bostonuniversityred] (y2) -- node[near end, anchor=south west, xshift=8pt]{$x_2$} ++(0, \framearrowlen);
        \draw[-latex, very thick, blue(pigment)] (y2) -- node[near end, anchor=north east]{$z_2$} ++(-\framearrowlen, 0);
        \node[draw, very thick, circle, greencross, ao(english)] (y3) at (\frameZeroPosX, \frameThreePosY) {};
        \draw[-latex, very thick, bostonuniversityred] (y3) -- node[near end, anchor=south west]{$x_3$} ++(0, \framearrowlen);
        \draw[-latex, very thick, blue(pigment)] (y3) --node[near end, anchor=north east]{$z_3$} ++(-\framearrowlen, 0);
        \node[draw, very thick, circle, bostonuniversityred, inner sep=1pt] (x4) at (\frameFourPosX, \frameThreePosY) {$\bullet$};
        \draw[-latex, very thick, ao(english)] (x4) -- node[near end, anchor=north west]{$y_4$} ++(\framearrowlen, 0);
        \draw[-latex, very thick, blue(pigment)] (x4) --node[near end, anchor=south west]{$z_4$} ++(0, \framearrowlen);
        \node[draw, very thick, circle, bostonuniversityred, inner sep=1pt] (x5) at (\frameFourPosX, \frameFivePosY) {$\bullet$};
        \draw[-latex, very thick, ao(english)] (x5) -- node[near end, anchor=south west]{$y_5$} ++(0, \framearrowlen);
        \draw[-latex, very thick, blue(pigment)] (x5) --node[near end, anchor=north east]{$z_5$} ++(-\framearrowlen, 0);
        \node[draw, very thick, circle, bostonuniversityred, inner sep=1pt] (x6) at (\frameSixPosX, \frameFivePosY) {$\bullet$};
        \draw[-latex, very thick, ao(english)] (x6) -- node[near end, anchor=south west]{$y_6$} ++(0, \framearrowlen);
        \draw[-latex, very thick, blue(pigment)] (x6) --node[near end, anchor=north east]{$z_6$} ++(-\framearrowlen, 0);
    \end{scope}
\end{scope}
\end{tikzpicture}

\caption{UR10 Kinematic Model}
\label{fig:ur10}

\end{figure}

Given the fundamental importance of an accurate dynamic characterization of manipulators, the identification of accurate dynamic models has been one of the major interests in robotics research for decades \cite{antonelli_systematic_1999}.
Traditional approaches involve modeling system dynamics using Newton-Euler equations \cite{gaz_identifying_2014} or Lagrangian methods \cite{hollerbach_recursive_1980}, and decomposing it to linearly separate kinematic variables from dynamic coefficients \cite{slotine_putting_1988}.

In the realm of system identification, dynamic coefficients are typically identified through online or offline techniques.
Online methods, such as composite adaptation and learning algorithms \cite{guo_composite_2023}, are often used in adaptive control to compensate for unknown dynamics and disturbances \cite{huang_composite_2021}.
However, offline estimation remains the prevalent strategy for identifying manipulator dynamics \cite{antonelli_systematic_1999, swevers_dynamic_2007}, which involves collecting data from specifically designed excitation trajectories \cite{luo_optimal_2023} and applying linear estimation algorithms \cite{atkeson_estimation_1986}.

Although these procedures have been performed on various motion-controlled manipulators \cite{gaz_extracting_2016, gaz_payload_2017, gaz_dynamic_2019}, two key challenges persist: linear models fail to fully capture the nonlinear dynamics due to friction effects, resulting in inaccuracies in torque prediction \cite{swevers_dynamic_2007, swevers_integrated_2000, al-bender_generalized_2005}; furthermore, unlike many commercial robots, industrial manipulators typically do not provide direct torque measurements, allowing access to joint currents only \cite{xu_robot_2022}.
This necessitates the estimation of \textit{motor drive gains}, which convert currents to torques, as part of the identification process.

This study focuses on the dynamic identification of the UR10 industrial manipulator, shown in Fig.~\ref{fig:ur10}, addressing the aforementioned challenges of estimating its linear dynamic coefficients, nonlinear joint friction currents, and motor drive gains.

\subsection{Contribution}

This paper's primary contribution is the comprehensive estimation of the UR10 robot dynamic coefficients, hence the provision of the complete dynamic model of this manipulator.
We compare our model to the state-of-the-art model \cite{gaz_model-based_2018} against two excitation trajectories.Our model consistently achieves lower estimation errors across all joints.
Unlike \cite{gaz_model-based_2018}, which solely estimates the static coefficients related to gravity and extracts other dynamic coefficients from a simulated model provided by the manufacturer, our approach estimates the full set of dynamic coefficients, executing excitation trajectories on the real hardware directly.
As mentioned in Section~\ref{sec:background}, our identification process consists of three stages:
\begin{enumerate*}[label=(\roman*)]
\item estimation of linear dynamic coefficients;
\item characterization of nonlinear friction currents;
\item computation of motor drive gains.
\end{enumerate*}

As an additional contribution, in a comparison with \cite{gaz_model-based_2018}, which performs a two-stage static estimation of motor drive gains, our method utilizes a one-shot algorithm based on dynamically exciting trajectories, yielding more accurate results.
For this, we adopt the methodology originally proposed in \cite{xu_robot_2022}, to map the whole set of estimated current-level dynamic parameters to the torque level, extending the identified model to include a generic payload at the manipulator's flange, building on formulations from \cite{gaz_payload_2017}.
To the best of the authors' knowledge, this is the first comprehensive study that includes motor drive gain estimation, dynamic identification, and payload integration for a single industrial manipulator.
Our findings, supported by a consistent investigation and analysis of all these features, are crucial for a complete and effective deployment of industrial robots.

Furthermore, we improve the practical applicability of our research providing a ROS2-ready software tool, the \textit{inverse dynamics solver} (IDS), that incorporates the identified dynamic model of the UR10 arm in an off-the-shelf library, providing users a simple tool to be straightforwardly integrated in control and planning applications.
Besides providing the manipulator dynamic model, the IDS is also designed to be easily reconfigurable, in that it allows users to include any known payload at the end-effector (EE) to obtain accurate joint torques for the entire arm-payload system, relieving users from the burden of performing a second estimation of the composite system \cite{atkeson_estimation_1986}.
The solver is experimentally validated on the UR10 with a known payload, using inertial parameters from the manufacturer's datasheet.

\subsection{Outline}

The rest of the article is structured as follows.
Section~\ref{sec:preliminaries} briefly describes the theoretical foundations on the identification problem, and introduces UR10's kinematic model.
Section~\ref{sec:dynamic-model-identification} presents the identification procedure we adopt to identify the robot's dynamic model.
Section~\ref{sec:experimental-results} reports the results obtained employing the procedure, comparing with the state-of-the-art model. 
Section~\ref{sec:conclusions} concludes the paper, summarizing the main findings of the present work and suggesting potential improvements left for future developments.
For convenience, the symbols and the notation adopted, as well as the dimensions of all the arrays utilized in the paper, are included in the \hyperref[sec:list-of-symbols]{List of Symbols} in appendix.

\section{Preliminaries}\label{sec:preliminaries}

\subsection{Manipulator Dynamic Model}\label{sec:manipulator-dynamic-model}

The manipulator dynamics in free-space can be modeled with the following equations of motion (EOM) \cite{siciliano_dynamics_2009}:
\begin{equation}\label{eq:eom}
    \gls{inertia}(\bm q) \ddot{\bm q} + \gls{coriolis}(\bm q, \dot{\bm q}) \dot{\bm q} + \gls{friction}(\dot{\bm q}) + \gls{gravity}(\bm q) = \bm\tau,
\end{equation}
where $\gls{inertia}(\bm q)$ is the inertia matrix, $\gls{coriolis}(\bm q, \dot{\bm q})$ is the matrix accounting for Coriolis and centrifugal effects, $\gls{friction}(\dot{\bm q})$ is the joint friction vector, $\gls{gravity}(\bm q)$ expresses the gravitational torques, $\bm\tau$ is the joint torque vector, and $\bm q, \dot{\bm q}, \ddot{\bm q}$ are the joint kinematic variables, namely positions, velocities and accelerations.

Joint torques $\bm\tau$ are strictly related to the currents $\bm v$ supplied to the joint motors, according to the following linear relationship \cite{hollerbach_model_2016}:
\begin{equation}\label{eq:motor-drive-gains}
    \bm\tau = \gls{mdg} \bm v, 
\end{equation}
where $\gls{mdg}$ is the diagonal matrix containing the joint motor drive gains.

According to the Lagrange formulation \cite{hollerbach_recursive_1980}, the components $\gls{inertia}$, $\gls{coriolis}$ and $\gls{gravity}$ are parametrized \wrt the link \textit{inertial parameters} including, for each link $i$, its mass $\gls{mass}_i$, center of mass (COM) vector $\gls{com}_i = [r_{i,x}, r_{i,y}, r_{i,z}]^\top$relative to the $i$-th kinematic frame, and the symmetric inertia tensor relative to the principal axes of inertia
\begin{equation}
    \gls{inertia-tensor-pai}_i = 
    \begin{bmatrix}
        \check I_{i,xx} & \check I_{i,xy} & \check  I_{i,xz}\\
        \check I_{i,xy} & \check I_{i,yy} & \check I_{i,yz}\\
        \check I_{i,xz} & \check I_{i,yz} & \check I_{i,zz}
    \end{bmatrix}.
\end{equation}

The inertial parameters can be grouped to constitute the robot \textit{dynamic parameters}
\begin{equation}\label{eq:dynamic-parameters}
\begin{split}
    \gls{d-parameters}_i = [& \gls{mass}_i,\;\;\gls{mass}_i r_{i,x},\;\;\gls{mass}_i r_{i,y},\;\;\gls{mass}_i r_{i,z},\\
    & I_{i,xx},\;\; I_{i,xy},\;\; I_{i,xz},\;\; I_{i,yy},\;\; I_{i,yz},\;\; I_{i,zz}]^\top,
\end{split}
\end{equation}
where $\gls{mass}_i \gls{com}_i$ is the first moment of inertia and $\gls{inertia-tensor}_i \triangleq \gls{inertia-tensor-pai}_i + \gls{mass}_i [\gls{com}_i]_\times^\top [\gls{com}_i]_\times$ represents the inertia tensor \wrt the origin of the $i$-th frame, in virtue of the Steiner theorem, with
\begin{equation}
    [\gls{com}_i]_\times \triangleq
    \begin{bmatrix}
         0 & -r_{i,z} & r_{i,y} \\
         r_{i,z} & 0 & -r_{i,x} \\
        -r_{i,y} & r_{i,x} & 0
    \end{bmatrix}
\end{equation}
being the skew-symmetric matrix operator applied to $\gls{com}_i$.

Ignoring the $\gls{friction}$ term, the EOM in \eqref{eq:eom} can be expressed in linear form, separating the dynamic parameters in \eqref{eq:dynamic-parameters} from kinematics-related expressions, as \cite{slotine_putting_1988}
\begin{equation}\label{eq:regressor}
    \gls{regressor}(\bm q, \dot{\bm q}, \ddot{\bm q}) \bm \pi = \bm\tau,
\end{equation}
where $\gls{regressor}(\bm q, \dot{\bm q}, \ddot{\bm q})$ is called \textit{regressor}, and $\gls{d-parameters} \triangleq [\gls{d-parameters}_1^\top, \ldots, \gls{d-parameters}_n^\top]^\top$ is the vector of the dynamic parameters for the whole $\gls{n-dof}$-link chain, with \gls{n-dof} being the number of degrees of freedom (DOF) of the robot.

\subsection{Joint Friction Model}\label{sec:joint-friction-model}

Usually, $\gls{friction}(\dot{\bm q})$ can be expressed as \cite{swevers_dynamic_2007}
\begin{equation}\label{eq:linear-friction-model}
    \gls{friction}(\dot{\bm q}) = \gls{f-o} + \gls{f-v} \dot{\bm q} + \gls{f-c} \mathbf{sgn}(\dot{\bm q}),
\end{equation}
where $\gls{f-o}, \gls{f-v}, \gls{f-c}$ are parameters representing the Coulomb friction offset, viscous friction and Coulomb friction coefficients, respectively, and multiplications are element-wise.
Since \eqref{eq:linear-friction-model} is linear \wrt \gls{f-o}, \gls{f-v} and \gls{f-c}, these parameters can be included in $\gls{d-parameters}$, thus being consistent with the linear model in \eqref{eq:regressor}.
Nevertheless, friction is notably a complex nonlinear phenomenon \cite{swevers_integrated_2000}, especially during motion reversal \cite{al-bender_generalized_2005}, hence a sigmoidal model is more appropriate \cite{gaz_model-based_2018}:
\begin{equation}\label{eq:friction-model}
    \gls{friction}(\dot{\bm q}) = \gls{f-o} + \gls{f-v} \dot{\bm q} + \frac{\gls{f-c}}{1 + \euler^{-\gls{f-mult}(\gls{f-add} + \dot{\bm q})}},
\end{equation}
with element-wise multiplications and exponentials.
Differently from \eqref{eq:linear-friction-model}, it is worth stressing that \eqref{eq:friction-model} presents a nonlinear dependency on the parameters parameters \gls{f-mult} and \gls{f-add}.

More specifically, friction exhibits a nonlinear behavior in a particular region (called \textit{nonlinearity region}) of the joint velocity domain, denoted with
\begin{equation}\label{eq:nonlinearity-region}
    \lvert \dot{\bm q} \rvert < \dot q^+ \; \Leftrightarrow \; \big\{ \dot{\bm q} \; \vert \; \exists j \in \{ 1, \ldots, n \}: \lvert \dot q_j \rvert < \dot q^+ \big\}.
\end{equation}
Henceforth, the short notation on the left-hand side will be used for convenience.
In this work we aim at comparing with \cite{gaz_model-based_2018}, thus we adopt the sigmoidal model \eqref{eq:friction-model}.
However, other formulations can be chosen in general: for instance, some models include the Stribeck effect \cite{tadese_passivity_2021} or the degressive friction \cite{ding_nonlinear_2018, weigand_dataset_2023}, whereas other approaches approximate the friction contribution with neural networks based on sigmoid-jump activation functions \cite{guo_composite_2019}.

\subsection{Modified Dynamic Coefficients Due to a Payload}\label{sec:modified-dynamic-coefficients-due-to-a-payload}

When a rigid payload is attached at the robot's EE, it introduces new dynamics that affect $\gls{d-parameters}_n$, according to the following modifications \cite{gaz_payload_2017}:
\begin{subequations}\label{eq:n-parameters}
\begin{align}
    \gls{mass}_n & \rightarrow \gls{mass}_n + \gls{mass}_L \label{eq:n-mass} \\
    \gls{com}_n & \rightarrow \frac{\gls{mass}_n \gls{com}_n + \gls{mass}_L \gls{com}_L}{\gls{mass}_n + \gls{mass}_L} \label{eq:n-com} \\
    \gls{inertia-tensor}_n & \rightarrow \gls{inertia-tensor}_n + \gls{inertia-tensor}_L \label{eq:n-it}
\end{align}
\end{subequations}
where $\gls{mass}_L$, $\gls{com}_L$ and $\gls{inertia-tensor}_L$ are the payload's mass, COM and inertia tensor \wrt \gls{n-dof}-th link frame, with
\begin{equation}\label{eq:payload-com}
    \tilde{\gls{com}}_L = \gls{p-transformation} \tilde{\gls{com}}_l, \qquad \tilde{\bm \bullet} \triangleq [\bm \bullet^\top, 1]^\top,
\end{equation}
and
\begin{equation}\label{eq:payload-it}
    \gls{inertia-tensor}_L = \gls{p-rotation} \gls{inertia-tensor}_l \bm R_l^n + \gls{mass}_L [\gls{com}_L]_\times^\top [\gls{com}_L]_\times.
\end{equation}

The linear transformations \eqref{eq:payload-com}--\eqref{eq:payload-it} are necessary to compute \eqref{eq:n-com}--\eqref{eq:n-it}, since the payload's COM $\gls{com}_l$ and inertia tensor $\gls{inertia-tensor}_l$ are typically expressed in frame $l$, \ie the payload origin frame, where the transform matrix $\gls{p-transformation}$, defined as
\begin{equation}
    \gls{p-transformation} \triangleq
    \begin{bmatrix}
        \gls{p-rotation} & \gls{p-translation} \\
        \bm 0_{1 \times 3} & 1
    \end{bmatrix},
\end{equation}
expresses the rotation \gls{p-rotation} and the translation \gls{p-translation} of $l$ \wrt the \gls{n-dof}-th link frame.

Because of \eqref{eq:n-mass}--\eqref{eq:n-com}, the first moment of inertia of the last link becomes \cite{gaz_payload_2017}
\begin{equation}
    \gls{mass}_n \gls{com}_n \rightarrow \gls{mass}_n \gls{com}_n + \gls{mass}_L \gls{com}_L,
\end{equation}
hence preserving the linearity property in \eqref{eq:regressor}, with
\begin{equation}\label{eq:payload-pi}
    \gls{d-parameters}_n \rightarrow \gls{d-parameters}_n + \gls{d-parameters}_L,
\end{equation}
where $\gls{d-parameters}_L$ indicates the payload's dynamic parameters, composed of $\gls{mass}_L$, $\gls{com}_L$ and $\gls{inertia-tensor}_L$ in \eqref{eq:dynamic-parameters}.

\subsection{UR10 Kinematic Model}

\begin{table}
\centering
\caption{UR10 DH parameters}
\label{tab:ur10}
\begin{tabular}{|c|c|c|c|c|} 
\hline
\textbf{Joint} & $\bm a~[\si{\milli\meter}]$ & $\bm\alpha~[\si{\radian}]$ & $\bm d~[\si{\milli\meter}]$ & $\bm\theta$ \\ 
\hline
Base & 0 & $-\nicefrac{\pi}{2}$ & 127.3 & $q_1$ \\
Shoulder & 612 & 0 & 0 & $q_2$ \\
Elbow & 572.3 & 0 & 0 & $q_3$ \\
Wrist 1 & 0 & $-\nicefrac{\pi}{2}$ & 163.941 & $q_4$ \\
Wrist 2 & 0 & $\nicefrac{\pi}{2}$ & 115.7 & $q_5$ \\
Wrist 3 & 0 & 0 & 92.2 & $q_6$ \\
\hline
\end{tabular}
\end{table}

Computing $\gls{regressor}(\bm q, \dot{\bm q}, \ddot{\bm q})$ requires the knowledge of the manipulator kinematics, usually devised applying the Denavit-Hartenberg (DH) convention on the dimensional properties given by the robot manufacturer \cite{antonelli_systematic_1999}.
However, since they can be subject to inaccuracies, one can mitigate the parameter uncertainty by performing a separate kinematic calibration procedure as a preliminary step \cite{sun_kinematic_2020}.
Nevertheless, in the context of dynamic model identification, these parameters are typically assumed to be known \cite{gaz_model-based_2018,antonelli_systematic_1999,gaz_identifying_2014,gaz_extracting_2016,gaz_payload_2017,gaz_dynamic_2019,xu_robot_2022}, and used as input to the estimation algorithm; therefore, the kinematic calibration problem falls outside the scope of this article.
As regards the UR10 robot, Fig.~\ref{fig:ur10} shows its kinematic model, also highlighting its joints and the relative frames, from which the DH parameters reported in Table~\ref{tab:ur10} are extracted\footnote{Link lengths available at the manufacturer website: \url{https://rb.gy/xxb5k5}}.

\section{Dynamic Model Identification}\label{sec:dynamic-model-identification}

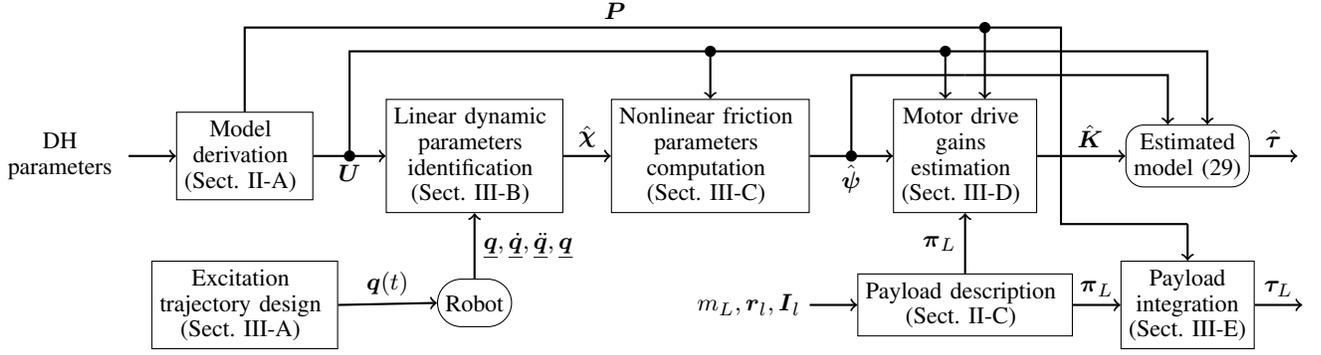
\begin{figure*}
\centering
\newcommand*{\myfontsize}{\small}
\myfontsize
\begin{tikzpicture}
\newcommand*{\edgeArrowHeight}{2em}
\newcommand*{\minBlockWidth}{5em}
\newcommand*{\maxBlockWidth}{8em}
\newcommand*{\varwidthBlock}[2][\maxBlockWidth]{%
    \begin{varwidth}{#1}
        \centering
        #2
    \end{varwidth}
}
\small
\matrix[matrix of nodes,
        nodes={rectangle, draw, text centered, minimum height=2em, minimum width=3em, font=\myfontsize},
        row sep=2em,
        column sep=2em,
        ] (m) 
{
    \varwidthBlock[6em]{Model derivation (Sect.~\ref{sec:manipulator-dynamic-model})} &
    \varwidthBlock{Linear dynamic parameters identification (Sect.~\ref{sec:linear-dynamic-parameters})} &
    \varwidthBlock{Nonlinear friction parameters computation (Sect.~\ref{sec:nonlinear-friction-parameters})} &
    \varwidthBlock[\minBlockWidth]{Motor drive gains estimation (Sect.~\ref{sec:motor-drive-gains})} &
    \node[draw, rounded corners=0.75em] (m-1-5) {\varwidthBlock[\minBlockWidth]{Estimated model \eqref{eq:estimated-torques-and-gains}}}; \\
    \varwidthBlock{Excitation trajectory design (Sect.~\ref{sec:excitation-trajectories})} &
    \node[draw, rounded corners=1em] (m-2-2){Robot}; & &
    \varwidthBlock{Payload description (Sect.~\ref{sec:modified-dynamic-coefficients-due-to-a-payload})} &
    \varwidthBlock[\minBlockWidth]{Payload integration (Sect.~\ref{sec:dynamic-model-identification-payload-integration})}\\
};

\tikzset{my-circle/.style={circle, fill=black, minimum size=1.5mm, inner sep=0pt}}

\draw[->, thick] ([xshift=-2em]m-1-1.west) -- node[pos=0,left,text width=5em,align=center] {DH parameters} (m-1-1.west);

\draw[->, thick] (m-1-1.east) -- (m-1-2.west) node[midway, below] {$\gls{c-minimal-regressor}$};

\path (m-1-1) -- (m-1-2) coordinate[pos=0.5] (midpoint-1-2);
\coordinate (m-1-3-above) at ([yshift=+2em]m-1-3.north);
\node[my-circle] at (midpoint-1-2) {};
\draw[->, thick] (midpoint-1-2) |- (m-1-3-above) -- (m-1-3.north) {};

\draw[->, thick] (m-1-2.east) -- (m-1-3.west) node[midway, above] {$\hat{\gls{c-d-coefficients}}$};

\draw[->, thick] (m-1-3.east) -- (m-1-4.west) node[midway, below] {$\hat{\gls{f-set}}$};

\path (m-1-3) -- (m-1-4) coordinate[pos=0.5] (midpoint-3-4);
\coordinate (m-1-4-above-1) at ([yshift=+1em]m-1-4.north);
\node[my-circle] at (midpoint-3-4) {};
\draw[-, thick] (midpoint-3-4) |- (m-1-4-above-1);
\draw[->, thick] (m-1-4-above-1) -| ([xshift=-0.83em]m-1-5.north);

\draw[->, thick] (m-1-4.east) -- (m-1-5.west) node[pos=0.6, above] {$\hat{\gls{mdg}}$};

\draw[->, thick] (m-1-5.east) -- node[midway,above] {$\hat{\bm\tau}$} ++(2em, 0);

\coordinate (m-1-4-above-3) at ([yshift=+3em]m-1-4.north);
\draw[->, thick]
(m-1-1.north) |-
([xshift=+0.83em]m-1-4-above-3) node[near end,above]{$\gls{regressor}$} --
([xshift=+0.83em]m-1-4.north);

\node[my-circle] at (m-1-3-above) {};
\draw[->, thick] (m-1-3-above) -| ([xshift=-0.83em]m-1-4.north);

\coordinate (m-1-4-above-2) at ([yshift=2em]m-1-4.north);
\node[my-circle] at ([xshift=-0.83em]m-1-4-above-2) {};
\draw[->, thick] ([xshift=-0.83em]m-1-4-above-2) -| ([xshift=+0.83em]m-1-5.north);

\path (m-1-5) -- (m-2-5) coordinate[pos=0.5] (aux);
\node[my-circle] at ([xshift=+0.83em]m-1-4-above-3) {};
\draw[-, thick] ([xshift=+0.83em]m-1-4-above-3) -| ([xshift=1em]m-1-4.east);
\draw[-, thick] ([xshift=1em]m-1-4.east) |- (aux);
\draw[->, thick] (aux) -- (m-2-5.north);

\draw[->, thick] ([xshift=-2em]m-2-4.west) -- node[pos=0,left] {$\gls{mass}_L, \gls{com}_l, \gls{inertia-tensor}_l$} (m-2-4.west);

\draw[->, thick] (m-2-1.east) -- node[midway,above] {$\bm q(t)$} (m-2-2.west);
\draw[->, thick] (m-2-2.north) -- node[midway,right] {$\underline{\bm q}, \underline{\dot{\bm q}}, \underline{\ddot{\bm q}}, \underline{\bm q}$} (m-1-2.south);
\draw[->, thick] (m-2-4.north) -- node[midway,left] {$\gls{d-parameters}_L$} (m-1-4.south);
\draw[->, thick] (m-2-4.east) -- node[midway,above]{$\gls{d-parameters}_L$} (m-2-5.west);
\draw[->, thick] (m-2-5.east) -- node[midway,above] {\gls{torques-p}} ++(2em, 0);

\end{tikzpicture}
\caption{Estimation procedure}
\label{fig:block-scheme}
\end{figure*}

This section describes the procedure we exercise to devise UR10's dynamic model, schematized in Figure~\ref{fig:block-scheme}.
As preliminary steps, we construct the robot's dynamic model (hence building the regressor) starting from its geometrical properties, and define the excitation trajectories to command in order to collect data used to identify the robot dynamics.
The actual estimation follows a three-stage approach, consisting in the following phases:
\begin{enumerate*}[label=(\roman*)]
    \item identification of linear dynamic parameters in \eqref{eq:regressor}, described in Section~\ref{sec:linear-dynamic-parameters};
    \item computation of nonlinear friction parameters in \eqref{eq:friction-model}, detailed in Section~\ref{sec:nonlinear-friction-parameters};
    \item estimation of motor drive gains in \eqref{eq:motor-drive-gains}, reported in Section~\ref{sec:motor-drive-gains}.
\end{enumerate*}
Lastly, in Section~\ref{sec:dynamic-model-identification-payload-integration} we describe how to extend the identified model to include the payload dynamics.

\subsection{Excitation Trajectories}\label{sec:excitation-trajectories}

As a preliminary step, a complete identification procedure requires collecting data to estimate and validate the dynamic parameters, which is typically done by commanding sinusoidal \textit{excitation trajectories}, parametrized as finite Fourier series \cite{swevers_dynamic_2007}
\begin{equation}\label{eq:excitation-trajectories}
    \bm q(t) = \bm q_0 + \sum_{k=1}^{\gls{n-fourier-components}}{\bm a_k \sin{\left( k \frac{2\pi}{T_f} t \right)} + \bm b_k\cos{\left( k \frac{2\pi}{T_f} t \right)}},
\end{equation}
where $\bm q_0$, $\bm a_k$ and $\bm b_k$ are Fourier coefficients, and $T_f$ is chosen to be a multiple of the robot's sampling period.

Once the trajectories are commanded, input/output data can be collected.
Let \gls{n-samples} be the number of recorded samples, and denote with $\underline\bullet$ the column-stack of a sequence of \gls{n-samples} data points.
Therefore, inputs $\underline{\bm q}, \underline{\dot{\bm q}}, \underline{\ddot{\bm q}}$ include measured kinematic states.
As regards the outputs, one can measure either joint torques $\underline{\bm\tau}$ or, as in the case of UR10, joint currents $\underline{\bm v}$.

The reader is invited to consider that designing excitation trajectories is paramount for accurate model estimation \cite{antonelli_systematic_1999}.
Therefore, it is worth stressing that a correct identification requires meeting the persistent excitation (PE) condition, \ie the reference signal \eqref{eq:excitation-trajectories} must include sufficiently rich spectral information to properly excite the robot dynamics \cite{guo_composite_2023, swevers_dynamic_2007}.
In practice, this typically requires trajectories of units to tens of thousands of samples \cite{antonelli_systematic_1999, xu_robot_2022, gaz_model-based_2018, gaz_identifying_2014, gaz_extracting_2016, gaz_dynamic_2019}, corresponding to execution times of, at most, units of minutes, assuming control frequencies in the hundreds to thousands of \si{\hertz}.
To this purpose, optimization of Fourier parameters has also been explored \cite{luo_optimal_2023}.

\subsection{Linear Dynamic Parameters}\label{sec:linear-dynamic-parameters}

As known, a subset of dynamic parameters in $\gls{d-parameters}$ can be estimated only in linear combinations \cite{atkeson_estimation_1986}, called \textit{dynamic coefficients}.
Hence, \eqref{eq:regressor} changes to
\begin{equation}\label{eq:minimal-regressor}
    \gls{minimal-regressor}(\bm q, \dot{\bm q}, \ddot{\bm q}) \gls{d-coefficients} = \bm\tau,
\end{equation}
where $\gls{d-coefficients}$ contains the dynamic coefficients, and $\gls{minimal-regressor}(\bm q, \dot{\bm q}, \ddot{\bm q})$ indicates the \textit{minimal regressor}, computed using Singular Value Decomposition (SVD) \cite{atkeson_estimation_1986, antonelli_systematic_1999}.

Retrieving $\gls{d-parameters}$ from $\gls{d-coefficients}$ requires a further optimization approach \cite{gaz_extracting_2016,gaz_dynamic_2019}, and goes beyond the scope of this article, since the computation of $\bm\tau$ from $\gls{d-coefficients}$ in \eqref{eq:minimal-regressor}, in place of $\gls{d-parameters}$ in \eqref{eq:regressor}, can equivalently be used to compute the EOM in \eqref{eq:eom}.

Dropping the dependencies on joint variables for the sake of brevity, $\gls{d-coefficients}$ can be estimated using either the Linear Least Square Estimation (LLSE) method \cite{atkeson_estimation_1986}
\begin{equation}\label{eq:llse}
    \hat{\gls{d-coefficients}} = \argmin_{\gls{d-coefficients}}{\lVert \underline{\bm\tau} - \underline{\gls{minimal-regressor}}\gls{d-coefficients} \rVert} = (\underline{\gls{minimal-regressor}}^\top \underline{\gls{minimal-regressor}})^{-1} \underline{\gls{minimal-regressor}}^\top \underline{\bm\tau},
\end{equation}
or the Weighted Least Square Estimation (WLSE) to be robust against inaccurate data \cite{swevers_dynamic_2007}
\begin{equation}\label{eq:wlse}
    \hat{\gls{d-coefficients}} = (\underline{\gls{minimal-regressor}}^\top \gls{weights} \underline{\gls{minimal-regressor}})^{-1} \underline{\gls{minimal-regressor}}^\top \gls{weights} \underline{\bm\tau},
\end{equation}
where $\gls{weights} = \diag{\{\nicefrac{1}{\gls{variance}}\}}_{k=1}^M$ is a diagonal matrix of weights, and $\gls{variance}$ is the variance of the $k$-th sample in $\underline{\bm\tau}$, with $(\bullet)^2$ indicating the element-wise square.
Plugging \eqref{eq:llse} or \eqref{eq:wlse} into \eqref{eq:minimal-regressor}, estimated joint torques can be retrieved for a given input state with
\begin{equation}\label{eq:estimated-torques}
    \hat{\bm\tau} = \gls{minimal-regressor}(\bm q, \dot{\bm q}, \ddot{\bm q}) \hat{\gls{d-coefficients}}.
\end{equation}

When torques cannot be directly measured, but currents are available instead, \eqref{eq:llse} and \eqref{eq:wlse} cannot be computed.
Hence, one can rewrite \eqref{eq:eom} in terms of joint currents using \eqref{eq:motor-drive-gains}, obtaining the following $n$ scalar equations:
\begin{equation}\label{eq:eom-currents}
	\frac{\gls{inertia-j}(\bm q)}{K_j} \ddot{\bm q} + \frac{\gls{coriolis-j}(\bm q, \dot{\bm q})}{K_j} \dot{\bm q} + \frac{f_j(q_j)}{K_j} + \frac{g_j(\bm q)}{K_j} = v_j,
\end{equation}
$\forall j \in \{ 1, \ldots, \gls{n-dof} \}$. Exploiting the same linearity property as in \eqref{eq:minimal-regressor}, \eqref{eq:eom-currents} becomes
\begin{equation}\label{eq:regressor-currents}
	\gls{minimal-regressor-j}(\bm q, \dot{\bm q}, \ddot{\bm q}) \gls{c-d-coefficients}_j = v_j, \quad \forall j \in \{ 1, \ldots, \gls{n-dof} \},
\end{equation}
with $\gls{c-d-coefficients}_j \triangleq \nicefrac{\gls{d-coefficients}}{K_j}$. 
In matrix form, \eqref{eq:regressor-currents} is rewritten as
\begin{equation}
	\begin{bmatrix}
		\bm y_1(\bm q, \dot{\bm q}, \ddot{\bm q}) & \cdots & \bm 0_{1 \times \gls{n-d-coefficients}} \\
		\vdots & \ddots & \vdots \\
		\bm 0_{1 \times \gls{n-d-coefficients}} & \cdots & {\bm y}_{\gls{n-dof}}(\bm q, \dot{\bm q}, \ddot{\bm q})
	\end{bmatrix} \gls{c-d-coefficients} = \bm v,
\end{equation}
where $\gls{c-d-coefficients} \triangleq [\gls{c-d-coefficients}_1^\top, \ldots, \gls{c-d-coefficients}_{\gls{n-dof}}^\top]^\top$ are the current-level dynamic coefficients.
Denoting the current-level minimal regressor as $\gls{c-minimal-regressor}(\bm q, \dot{\bm q}, \ddot{\bm q}) \triangleq \blockdiag{ \{ \gls{minimal-regressor-j}(\bm q, \dot{\bm q}, \ddot{\bm q}) \} }_{j=1}^{\gls{n-dof}}$, \eqref{eq:wlse} becomes
\begin{equation}\label{eq:wlse-currents}
	\hat{\gls{c-d-coefficients}} = (\underline{\gls{c-minimal-regressor}}^\top \gls{weights} \underline{\gls{c-minimal-regressor}})^{-1} \underline{\gls{c-minimal-regressor}}^\top \gls{weights} \underline{\bm v},
\end{equation}
and \eqref{eq:estimated-torques} translates into
\begin{equation}\label{eq:estimated-currents}
	\hat{\bm v} = \gls{c-minimal-regressor}(\bm q, \dot{\bm q}, \ddot{\bm q}) \hat{\gls{c-d-coefficients}}.
\end{equation}

To summarize, the first step of the identification procedure consists in computing the current-level minimal regressor \gls{c-minimal-regressor} from the torque-level equivalent \gls{minimal-regressor}, and estimating current-level dynamic coefficients \gls{c-d-coefficients} after collecting data $(\bm q, \dot{\bm q}, \ddot{\bm q}, \bm v)$: estimated currents $\hat{\bm v}$ are finally computed with \eqref{eq:estimated-currents}.
In the light of the discussion in Section~\ref{sec:joint-friction-model}, one should consider that \eqref{eq:estimated-currents} holds if and only if the complement of \eqref{eq:nonlinearity-region} is satisfied, \ie $\lvert \dot{\bm q} \rvert > \dot q^+$, \ie where the joint friction currents behave linearly according to \eqref{eq:linear-friction-model}.

\subsection{Nonlinear Friction Parameters}\label{sec:nonlinear-friction-parameters}

In the second stage of our estimation procedure, we estimate the set of nonlinear friction parameters, denoted with $\gls{f-set} \triangleq \bigcup_{j=1}^n{{\gls{f-set}}_j}$, where ${\gls{f-set}}_j \triangleq \{ f_{o,j}, f_{v,j}, f_{c,j}, \delta_j, \nu_j \}$.
Given the dynamic coefficients identified in \eqref{eq:wlse-currents}, we compute the estimated friction currents as
\begin{equation}\label{eq:estimated-friction-currents}
	\hat{\bm v}_f(\dot{\bm q}) = \bm v - \gls{c-minimal-regressor-f}(\bm q, \dot{\bm q}, \ddot{\bm q}) \hat{\gls{c-d-coefficients}}_f,
\end{equation}
where $\gls{c-minimal-regressor-f}$ (\resp $\hat{\gls{c-d-coefficients}}_f$) includes the columns (\resp rows) of $\gls{c-minimal-regressor}$ (\resp $\hat{\gls{c-d-coefficients}}$), excluding the linear friction parameters $\gls{f-o}, \gls{f-v}, \gls{f-c}$.

Given a set of parameters $\gls{f-set}$, we denote with $\gls{torques-f-set}(\dot{\bm q})$ the joint friction torques in \eqref{eq:friction-model} evaluated with $\gls{f-set}$; hence, we define the corresponding joint friction currents as $\gls{c-f-set}(\dot{\bm q}) \triangleq \gls{mdg}^{-1} \gls{torques-f-set}(\dot{\bm q})$.
Finally, we estimate the friction parameters by solving the following nonlinear optimization problem:
\begin{equation}\label{eq:estimated-friction-parameters}
	\hat{\gls{f-set}} = \argmin_{\gls{f-set}}{ \left\{ \hat{\underline{\bm v}}_f(\dot{\bm q}) - \underline{\bm v}_{\gls{f-set}}(\dot{\bm q}) \right\} },
\end{equation}
with $\hat{\underline{\bm v}}_f(\dot{\bm q})$ and $\underline{\bm v}_{\gls{f-set}}(\dot{\bm q})$ representing, in this case, the column-stack of $\hat{\bm v}_f$ and $\gls{c-f-set}$ evaluated on the samples \ST $\lvert \dot{\bm q} \rvert < \dot q^+$ holds, thus exhibiting the sigmoidal characteristics in \eqref{eq:friction-model}.

Notably, \eqref{eq:estimated-friction-parameters} replaces the linear friction parameters estimated in \eqref{eq:wlse-currents}, fitting them to the nonlinear formulation.
Finally, the estimated currents are
\begin{equation}\label{eq:estimated-currents-with-friction}
	\hat{\bm v} = \gls{c-minimal-regressor-f}(\bm q, \dot{\bm q}, \ddot{\bm q}) \hat{\gls{c-d-coefficients}}_f + \bm v_{\hat{\gls{f-set}}}(\dot{\bm q}).
\end{equation}

Readers should be aware that joint friction estimation for industrial robots is currently an active research topic \cite{ding_nonlinear_2018}.
Indeed, solving the nonlinear problem \eqref{eq:estimated-friction-parameters} with a seed-dependent algorithm does not guarantee a globally optimal solution: this well-known issue is typical of gradient-based solvers (also adopted by, \eg, \cite{gaz_model-based_2018,xu_robot_2022}).
Exploring the most suitable friction model and resolution algorithm falls outside the scope of this article, since this particular sub-problem requires further specific investigation.

\subsection{Motor Drive Gains} \label{sec:motor-drive-gains}

Once the currents are estimated with \eqref{eq:estimated-currents-with-friction}, the third and final stage of our estimation procedure is performed, \ie, $\hat{\bm v}$ are converted into $\hat{\bm\tau}$.
According to \eqref{eq:motor-drive-gains}, once the motor drive gains $\hat{\gls{mdg}}$ are estimated, the following relationship holds:
\begin{equation}\label{eq:estimated-torques-and-gains}
	\hat{\bm\tau} = \hat{\gls{mdg}} \hat{\bm v}.
\end{equation}
In this work, we replicate the procedure performed by \cite{xu_robot_2022} on our manipulator to compute $\hat{\gls{mdg}}$.
This section reports and clarifies its steps, for the sake of completeness.

First, the algorithm requires the execution of a set of trajectories run with the manipulator arm only (let us denote this scenario with $a$), and a set of trajectories run with the robot holding a payload (scenario $b$).
After the trajectories are executed, \gls{n-samples} samples are collected, with $\gls{n-samples} \triangleq \gls{n-samples}^a + \gls{n-samples}^b$.
As regards the payload dynamic parameters $\gls{d-parameters}_L$, it is assumed that only a certain number \gls{n-u-p-d-parameters} of them are unknown.
This is a reasonable assumption since, for instance, one can simply use a scale to obtain the payload mass $\gls{mass}_L$ at least, thus $\gls{n-u-p-d-parameters} \le 9$ is assumed to hold in general.
Hence, $\gls{d-parameters}_L$ can be partitioned into $\gls{u-p-d-parameters}$ and $\gls{k-p-d-parameters}$, \ie, the sets of unknown and known payload dynamic parameters, respectively.

Consider the following equation, holding for the $j$-th joint:
\begin{equation}\label{eq:linear-currents-with-payload}
	\begin{bmatrix}
		\underline{\bm v}_j^a - \underline{\bm v}_{\hat{\gls{f-set}},j}^a \\
		\underline{\bm v}_j^b - \underline{\bm v}_{\hat{\gls{f-set}},j}^b
	\end{bmatrix} =
	\begin{bmatrix}
		\underline{\bm u}_{f,j}^a & \bm 0_{\gls{n-samples}^a \times \gls{n-u-p-d-parameters}} & \bm 0_{\gls{n-samples}^a \times 1} \\
		\underline{\bm u}_{f,j}^b & \underline{\bm p}_{uL,j} & \underline{\bm p}_{kL,j} \gls{k-p-d-parameters}
	\end{bmatrix}
	\begin{bmatrix}
		\gls{c-d-coefficients}_{f,j} \\
		\frac{\gls{u-p-d-parameters}}{K_j} \\
		\frac{1}{K_j}
	\end{bmatrix},
\end{equation}
simplified as
\begin{equation}
	\underline{\bm v}_j - \underline{\bm v}_{\hat{\gls{f-set}},j} = \gls{regressor-mdg-j} \gls{vector-mdg-j},
\end{equation}
where $\underline{\bm v}_j - \underline{\bm v}_{\hat{\bm\psi},j} \triangleq [ ( \underline{\bm v}_j^a - \underline{\bm v}_{\hat{\bm\psi},j}^a )^\top, ( \underline{\bm v}_j^b - \underline{\bm v}_{\hat{\bm\psi},j}^b )^\top ]^\top$ stacks the \textit{linear} current samples collected in scenarios $a$ and $b$ respectively, $\gls{c-d-coefficients}_{f,j}$ are the rows of $\gls{c-d-coefficients}_j$ excluding linear friction parameters, $\underline{\bm u}_{f,j}^a$ and $\underline{\bm u}_{f,j}^b$ are the column-stacks of $\gls{c-minimal-regressor-f-j}$ (\ie, the $j$-th row of $\gls{c-minimal-regressor-f}$) evaluated on kinematic states measured in scenarios $a$ and $b$ respectively, and $\underline{\bm p}_{uL,j}$ (\resp $\underline{\bm p}_{kL,j}$) is a column-stack of \gls{regressor-u-p-j} (\resp \gls{regressor-k-p-j}), namely the $j$-th row of \gls{regressor-u-p} (\resp \gls{regressor-k-p}), \ie the regressor \wrt \gls{u-p-d-parameters} (\resp \gls{k-p-d-parameters}), evaluated on input samples retrieved in scenario $b$.
Since \eqref{eq:linear-currents-with-payload} uses a linear model, for the sake of clarity we highlight that henceforth all the kinematics-dependent variables are considered to be evaluated on samples in the linearity region only.

Similarly to \eqref{eq:wlse}, $\forall j \in \{ 1, \ldots, \gls{n-dof} \}$, \gls{regressor-mdg-j} is used to estimate \gls{vector-mdg-j} with WLSE:
\begin{equation}\label{eq:wlse-payload}
	\hat{\bm\zeta}_j = (\regressormdg_j^\top \gls{weights} \gls{regressor-mdg-j} )^{-1} \regressormdg_j^\top \gls{weights} \underline{\bm v}_j.
\end{equation}
Then, $\hat K_j$ can be simply extracted from $\hat{\bm\zeta}_j$ with
\begin{equation}
	\hat K_j = \frac{1}{\gls{decision-vector-j} \hat{\bm\zeta}_j}, \qquad \gls{decision-vector-j} \triangleq [\bm 0_{1 \times (\gls{n-d-coefficients} + \gls{n-u-p-d-parameters} - 3)}, \; 1].
\end{equation}

However, there might exist a joint $j^*$ for which the matrices $\{ \gls{regressor-mdg-j} \}_{j^*}^{\gls{n-dof}}$ are rank-deficient; thus, $\{ K_j \}_{j^*}^{\gls{n-dof}}$ cannot be directly estimated, as \gls{vector-mdg-j} combines linearly with other parameters.
In these cases, \gls{vector-mdg-j} must be partitioned into two parts:
\begin{equation}
    \gls{vector-mdg-j} = [\identifiableparameters_j^\top, \unidentifiableparameters_j^\top]^\top,
\end{equation}
where \gls{identifiable-parameters-j} and \gls{unidentifiable-parameters-j} represent the identifiable and unidentifiable parameters, respectively.
Denoting the size of \gls{unidentifiable-parameters-j} with $\lvert \gls{unidentifiable-parameters-j} \rvert$, the following equation holds, according to the QR decomposition of \gls{regressor-mdg-j}:
\begin{align}    
    \gls{regressor-mdg-j} \gls{vector-mdg-j} &= [\gls{regressor-mdg-j-li}, \gls{regressor-mdg-j-ld}]
    \begin{bmatrix}
        \gls{identifiable-parameters-j} \\
        \gls{unidentifiable-parameters-j}
    \end{bmatrix} = \nonumber \\
    &= [\gls{regressor-mdg-j-li}, \gls{regressor-mdg-j-ld}]
    \begin{bmatrix}
        \gls{regrouping-vector-j} \\
        \bm 0_{\lvert \gls{unidentifiable-parameters-j} \rvert \times 1}
    \end{bmatrix} = \gls{regressor-mdg-j-li} \gls{regrouping-vector-j} = \\
    &= [\bm{\mathcal Q}_{j,\phi}, \bm{\mathcal Q}_{j,\varphi}]
    \begin{bmatrix}
        \gls{r-mdg-j-li} & \gls{r-mdg-j-ld} \\
        \bm 0_{\gls{n-samples} \times \gls{n-identifiable-parameters-j}} & \bm 0_{\gls{n-samples} \times \lvert \gls{unidentifiable-parameters-j} \rvert}
    \end{bmatrix}
    \begin{bmatrix}
        \gls{regrouping-vector-j} \\
        \bm 0_{\lvert \gls{unidentifiable-parameters-j} \rvert \times 1}
    \end{bmatrix}, \nonumber
\end{align}
where $\gls{regrouping-vector-j} \triangleq \gls{identifiable-parameters-j} + \rmdg_{j,\phi}^\dagger \gls{r-mdg-j-ld}\gls{unidentifiable-parameters-j}$ is the regrouping vector of \gls{vector-mdg-j}; \gls{regressor-mdg-j-li} and \gls{regressor-mdg-j-ld} contain the linearly independent and dependent columns of \gls{regressor-mdg-j}, respectively, and $\bm{\mathcal Q}_{j,\phi}$, $\bm{\mathcal Q}_{j,\varphi}$, \gls{r-mdg-j-li} and \gls{r-mdg-j-ld} are, respectively, the orthogonal and upper triangular matrices of \gls{regressor-mdg-j-li} and \gls{regressor-mdg-j-ld}, with $\bullet^\dagger$ indicating the Moore-Penrose pseudo-inverse.

After the QR factorization of \gls{regressor-mdg-j}, the following optimization problem must be solved to extract $\hat K_j$:
\begin{subequations}\label{eq:estimated-motor-drive-gains}
\begin{align}
    \hat{\bm\zeta}_j = & \argmin_{\gls{vector-mdg-j}}{ \lVert \gls{regressor-mdg-j-li} \gls{regrouping-vector-j} - (\underline{\bm v}_j - \underline{\bm v}_{\hat{\gls{f-set}},j}) \rVert } \\
    \st \quad & K_j = \frac{1}{\gls{decision-vector-j} \gls{vector-mdg-j} }, \\
    & K_j^- < K_j < K_j^+,
\end{align}
\end{subequations}
where $K_j^-$ and $K_j^+$ are minimum and maximum bounds to impose on $K_j$.

\subsection{Payload Integration}\label{sec:dynamic-model-identification-payload-integration}

Rewriting \eqref{eq:regressor} to separate the $n$-th link, one has:
\begin{equation}\label{eq:regressor-separated}
    \gls{torques-arm} = \gls{regressor} \gls{d-parameters} = \gls{regressor-1-to-nm1} \gls{d-parameters-1-to-nm1} + \gls{regressor-n} \gls{d-parameters}_{\gls{n-dof}},
\end{equation}
where \gls{regressor-1-to-nm1} represents the columns of the regressor matrix related to the dynamic parameters of links from $1$ to $\gls{n-dof}-1$, \gls{regressor-n} contains the columns related to link \gls{n-dof} only, and $\gls{d-parameters-1-to-nm1} \triangleq [\gls{d-parameters}_1^\top, \ldots, \gls{d-parameters}_{\gls{n-dof}-1}^\top]^\top$.

In virtue of \eqref{eq:payload-pi}, in the presence of a payload, \eqref{eq:regressor-separated} changes to
\begin{equation}\label{eq:tau-arm-tau-l}
\begin{split}
    \bm\tau = \gls{regressor} \gls{d-parameters} &= \gls{regressor-1-to-nm1} \gls{d-parameters-1-to-nm1} + \gls{regressor-n} (\gls{d-parameters}_n + \gls{d-parameters}_L) = \\
    &= \underbrace{\gls{regressor-1-to-nm1} \gls{d-parameters-1-to-nm1} + \gls{regressor-n} \gls{d-parameters}_{\gls{n-dof}}}_{=\gls{torques-arm}} + \gls{regressor-n} \gls{d-parameters}_L = \\
    &= \gls{torques-arm} + \gls{regressor-n} \gls{d-parameters}_L = \gls{torques-arm} + \gls{torques-p},
\end{split}
\end{equation}
where \gls{torques-arm} are the torques due to the arm motion only and \gls{torques-p} are those due to the payload.
Therefore, once \gls{torques-arm} is estimated and $\gls{d-parameters}_L$ is known, one can obtain the joint torques of the composed arm-payload system $\bm\tau$ simply using the regressor \gls{regressor-n} (which only contains kinematic information) and $\gls{d-parameters}_L$.
The latter is typically known from the datasheet, as the one that will be used in Section~\ref{sec:experimental-results-payload-integration}, or can be estimated separately, as in \cite{gaz_payload_2017}.

\section{Experimental Results}\label{sec:experimental-results}

In this section, we report the results of the estimation of the dynamic coefficients of the UR10 robot, shown in Fig.~\ref{fig:ur10}.

\subsection{Task Setup \& Materials}

\subsubsection{Manipulator}

To acquire input/output data from the 6-DOF UR10 manipulator, we use a Python library provided by the manufacturer\footnote{\url{https://rb.gy/dnuvs9}}, from which we can measure $\bm q$, $\dot{\bm q}$ and $\bm v$, whereas $\ddot{\bm q}$ can be computed using the backward Euler approximation from $\dot{\bm q}$.
To cope with the inherent noise of $\ddot{\bm q}$ \cite{guo_locally_2022}, due to the inescapable measurement noise of $\dot{\bm q}$ further deteriorated by discrete differentiation, the time history of the acquired samples is filtered out of high-frequency noise with a low-pass filter, implemented in MATLAB using the \texttt{designfilt} function.
Excitation trajectories in the shape of \eqref{eq:excitation-trajectories} are commanded using a \texttt{ros\_control} \cite{chitta_ros_control_2017} position-based joint trajectory controller, through the ROS driver developed by UR\footnote{\url{https://rb.gy/nhd821}}.
Two of the excitation trajectories we defined are used as validation set: henceforth, they will be referred to as Trajectory A and Trajectory B.

\subsubsection{Payload}

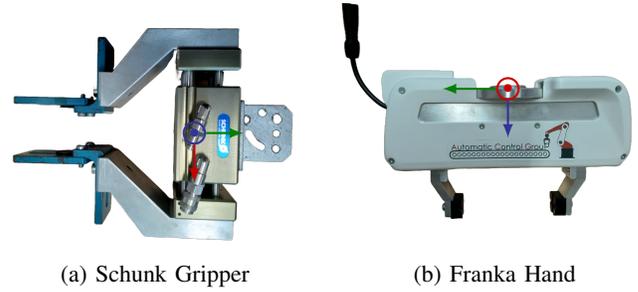
\begin{figure}
    \centering

    \begin{subfigure}[b]{0.49\columnwidth}
        \centering
        \begin{tikzpicture}
        \node[anchor=south west, inner sep=0] (image) at (0,0){\includegraphics[width=\textwidth]{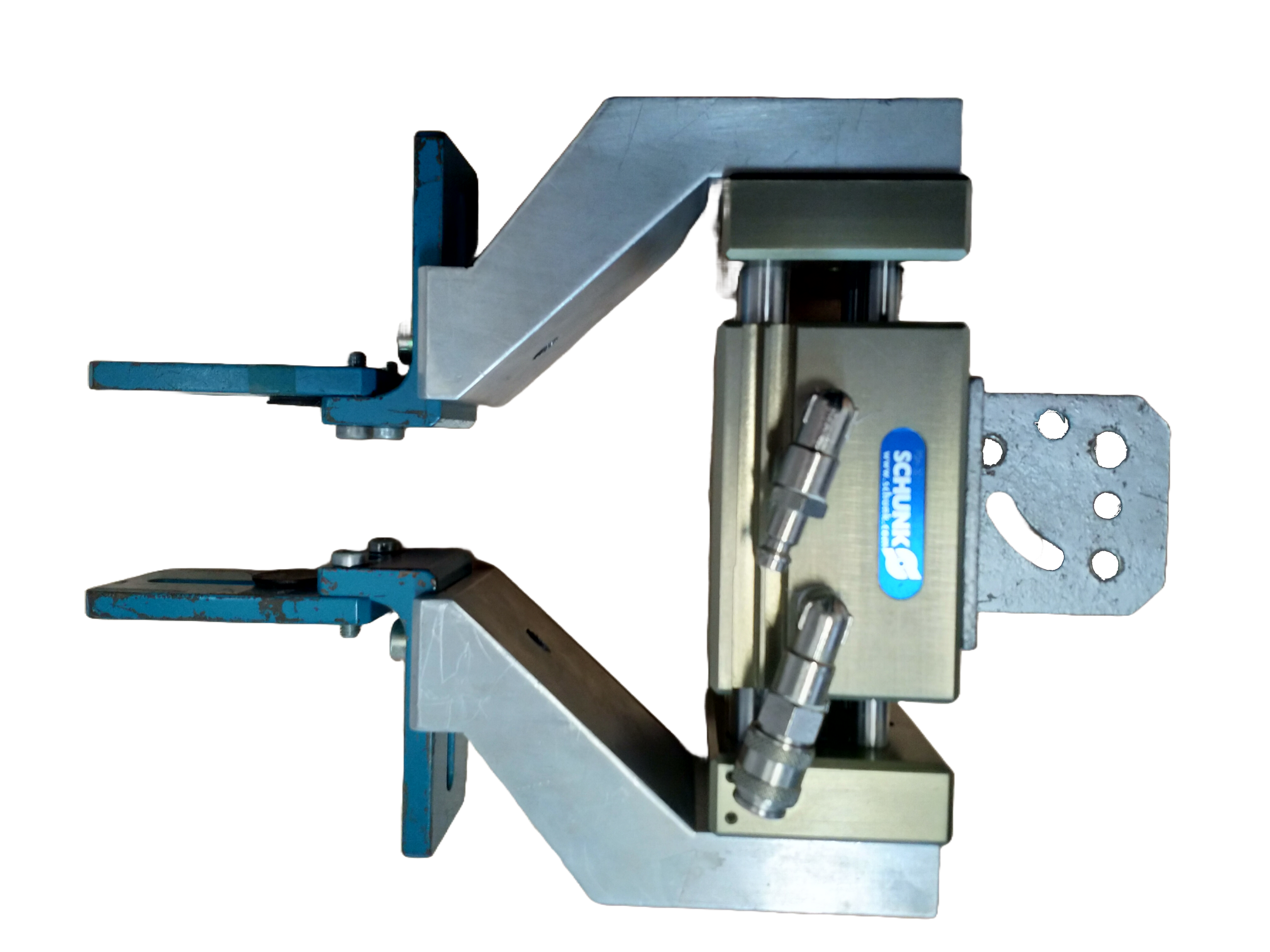}};
        \begin{scope}[x={(image.south east)},y={(image.north west)}]
            \node[draw, thick, circle, blue(pigment), inner sep=1pt, font=\scriptsize] (xg) at (0.62, 0.47) {$\bullet$};
            \draw[-latex, thick, ao(english)] (xg) -- ++(0.15, 0.0);
            \draw[-latex, thick, bostonuniversityred] (xg) -- ++(0.0, -0.2);
        \end{scope}
        \end{tikzpicture}
        \caption{Schunk Gripper}
        \label{fig:gripper}
    \end{subfigure}
    \hfill
    \begin{subfigure}[b]{0.49\columnwidth}
        \centering
        \begin{tikzpicture}
        \node[anchor=south west, inner sep=0] (image) at (0,0){\includegraphics[width=\textwidth]{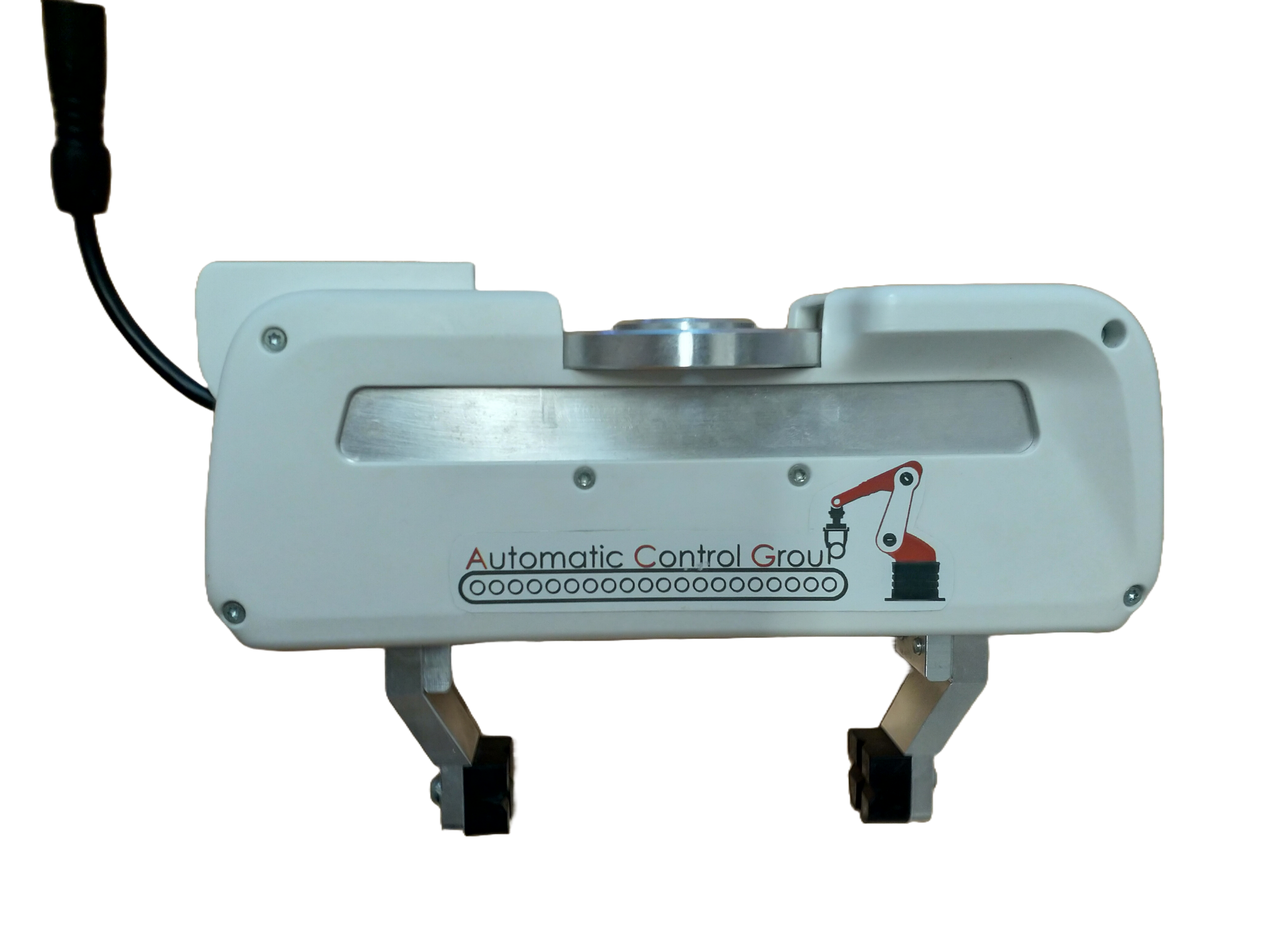}};
        \begin{scope}[x={(image.south east)},y={(image.north west)}]
            \node[draw, thick, circle, bostonuniversityred, inner sep=1pt, font=\scriptsize] (xh) at (0.54, 0.65) {$\bullet$};
            \draw[-latex, thick, ao(english)] (xh) -- ++(-0.2, 0.0);
            \draw[-latex, thick, blue(pigment)] (xh) -- ++(0.0, -0.2);
        \end{scope}
        \end{tikzpicture}
        \caption{Franka Hand}
        \label{fig:franka-hand}
    \end{subfigure}

    \caption{Payloads}
    \label{fig:payloads}
\end{figure}

\begin{table}
\centering
\caption{Franka Hand inertial parameters}
\label{tab:franka-hand}
\begin{tabular}{|c|ccc|ccc|} 
\hline
\textbf{Mass [\si{\kilo\gram}]} & \multicolumn{3}{c|}{\textbf{COM [\si{\milli\meter}]}} & \multicolumn{3}{c|}{\textbf{Inertia matrix [\si{\gram\meter\squared}]}} \\ 
\hline
$m$ & $c_x$ & $c_y$ & $c_z$ & $I_{xx}$ & $I_{yy}$ & $I_{zz}$ \\
0.73 & 0 & 10 & 30 & 1 & 2.5 & 1.7 \\
\hline
\end{tabular}
\end{table}

In our experiments we use two different payloads (Fig.~\ref{fig:payloads}): a Schunk PSH-32/1 gripper (Fig.~\ref{fig:gripper}) to estimate the motor gains according to the procedure defined in Section~\ref{sec:motor-drive-gains}, and the Franka Hand (Fig.~\ref{fig:franka-hand}) to validate our configurable IDS.
As regards the former, we only know its mass (hence $\gls{n-u-p-d-parameters} = 9$) by weighing, \ie $\gls{mass}_L = \SI{4.823}{\kilo\gram}$, while the inertial parameters of the latter are known from the datasheet\footnote{\url{https://rb.gy/l8wb9y}} and reported in Table~\ref{tab:franka-hand}.

\subsubsection{Software}\label{sec:software}

Our UR10 IDS exposes interfaces, written in C++, to extract $\gls{inertia}(\bm q)$, $\gls{coriolis}(\bm q, \dot{\bm q})\dot{\bm q}$, $\gls{friction}(\dot{\bm q})$, $\gls{gravity}(\bm q)$ or $\bm\tau$.
The payload inertial parameters can be added in a separate configuration file, without any change to the software itself, thanks to the implementation of \eqref{eq:tau-arm-tau-l} separating \gls{torques-arm} and \gls{torques-p}. 
Readers are invited to visit our public repository\footnote{\url{https://codeocean.com/capsule/8515919/tree/v2}} and consult the related documentation to easily get acquainted with the solver and perform demos and unit tests.
Further details on the IDS are given in Section~\ref{sec:software-implementation}.
The estimations \eqref{eq:wlse-currents}, \eqref{eq:wlse-payload} are implemented using the \texttt{robustfit} function, which automatically computes $\gls{weights}$ and applies WLSE.
The optimization problem \eqref{eq:estimated-friction-parameters} is solved with \texttt{fminunc} setting $\dot q^+ \coloneqq \SI{0.17}{\radian\per\second}$, whereas \eqref{eq:estimated-motor-drive-gains} is solved with \texttt{lsqlin}, as it is a LLSE problem, choosing $K_j^- \coloneqq \SI{10}{\newton\meter\per\ampere}$ and $K_j^+ \coloneqq \max_{i \in \{ 1, \ldots, j - 1 \}}{K_i} $.

\subsubsection{Performance indices}
With
\begin{equation}
    \MSE(\bm x, \bm y) = \frac{1}{\varsigma}\sum_{k=1}^{\varsigma}{(x_k - y_k)^2}
\end{equation}
we denote the Mean Squared Error (MSE) between two generic vectors $\bm x, \bm y \in \mathbb R^\varsigma$, and with
\begin{equation}\label{eq:mnae}
    \MNAE(\bm x, \bm y) = \frac{200}{\varsigma}\sum_{k=1}^{\varsigma}{\frac{\lvert x_k - y_k \rvert}{\max_{k}{x_k} - \min_{k}{x_k}}}
\end{equation}
we define the Mean Normalized Absolute Error (MNAE) between $\bm x$ and $\bm y$, \ie the percentage of how much $\bm y$ deviates from $\bm x$, normalized against the range of the samples in $\bm x$.
The index \eqref{eq:mnae} is particularly useful in fairly assessing our estimated currents accuracy, \eg with $\MNAE(\bm v_j, \hat{\bm v}_j)$, in that the estimation error is weighted differently depending on the actual current domain.

\subsection{Dynamic Model Estimation}

\subsubsection{Joint Currents}

\begin{figure*}
    \centering
    \newcommand*{\figsize}{0.32\textwidth}

    \begin{center}
    \footnotesize{{\legblue} measured \hspace{0.5em} {\boxgray} nonlinearity region \hspace{0.5em} {\legred} estimated with \cite{gaz_model-based_2018} \hspace{0.5em} {\legyellow} our estimation}
    \end{center}

    \begin{subfigure}[b]{\figsize}
        \centering
        \includegraphics[width=\textwidth]{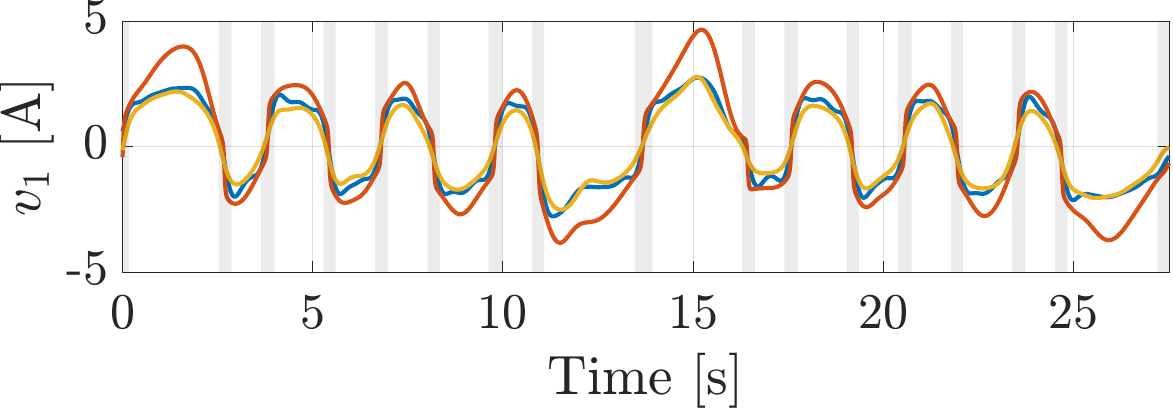}
    \end{subfigure}
    \hfill
    \begin{subfigure}[b]{\figsize}
        \centering
        \includegraphics[width=\textwidth]{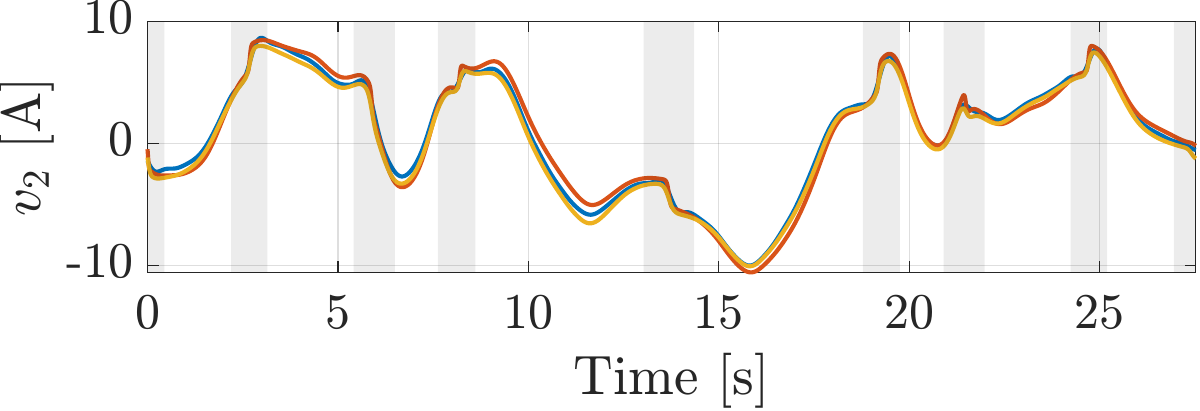}
    \end{subfigure}
    \hfill
    \begin{subfigure}[b]{\figsize}
        \centering
        \includegraphics[width=\textwidth]{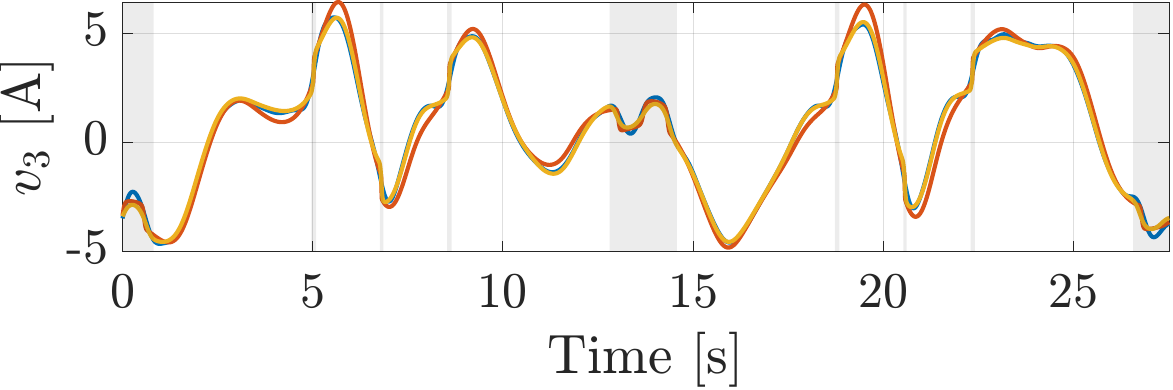}
    \end{subfigure}
    \\ \vspace{0.25em}
    \begin{subfigure}[b]{\figsize}
        \centering
        \includegraphics[width=\textwidth]{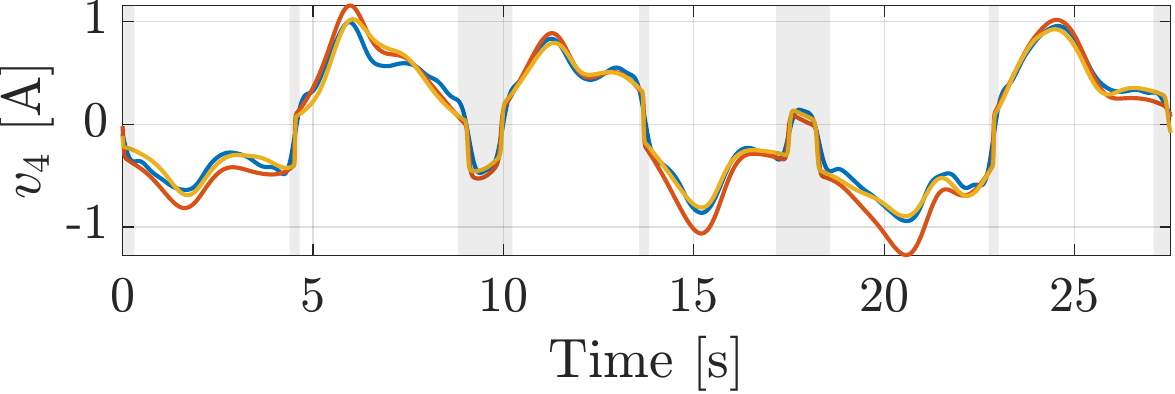}
    \end{subfigure}
    \hfill
    \begin{subfigure}[b]{\figsize}
        \centering
        \includegraphics[width=\textwidth]{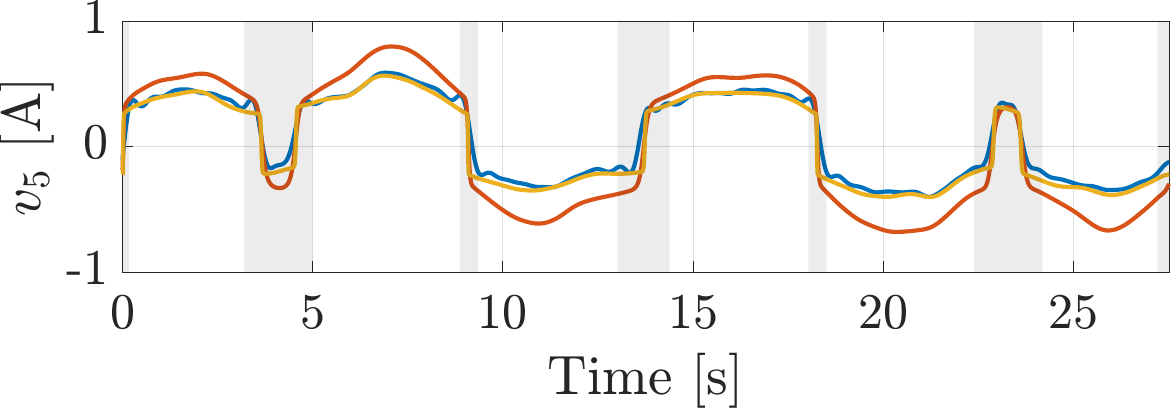}
    \end{subfigure}
    \hfill
    \begin{subfigure}[b]{\figsize}
        \centering
        \includegraphics[width=\textwidth]{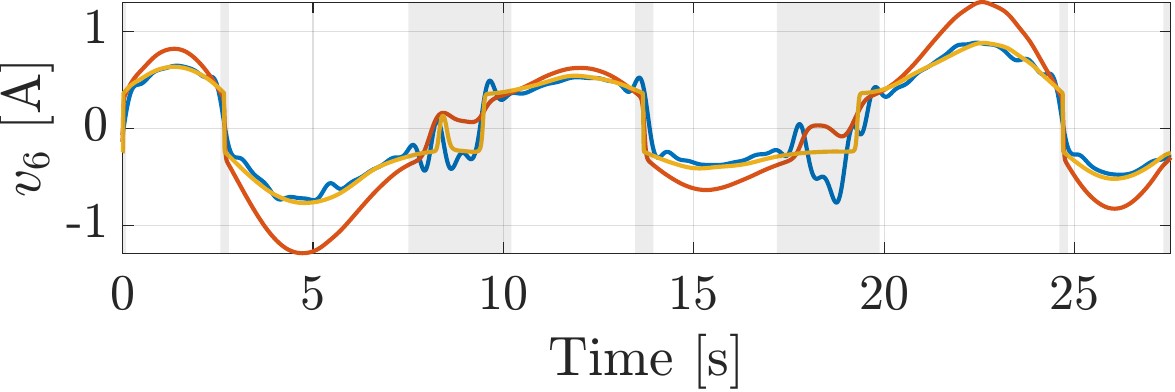}
    \end{subfigure}
    \\ \vspace{0.25em}  
    \begin{subfigure}[b]{\figsize}
        \centering
        \includegraphics[width=\textwidth]{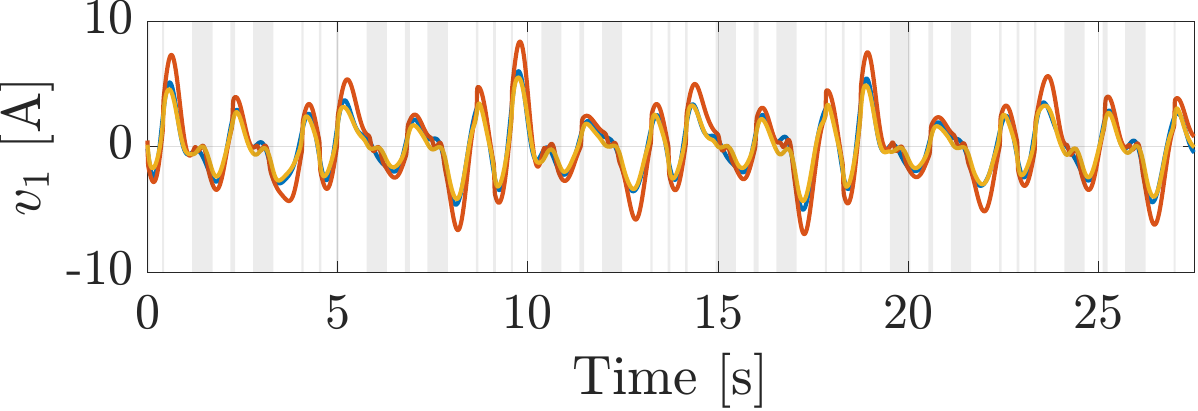}
    \end{subfigure}
    \hfill
    \begin{subfigure}[b]{\figsize}
        \centering
        \includegraphics[width=\textwidth]{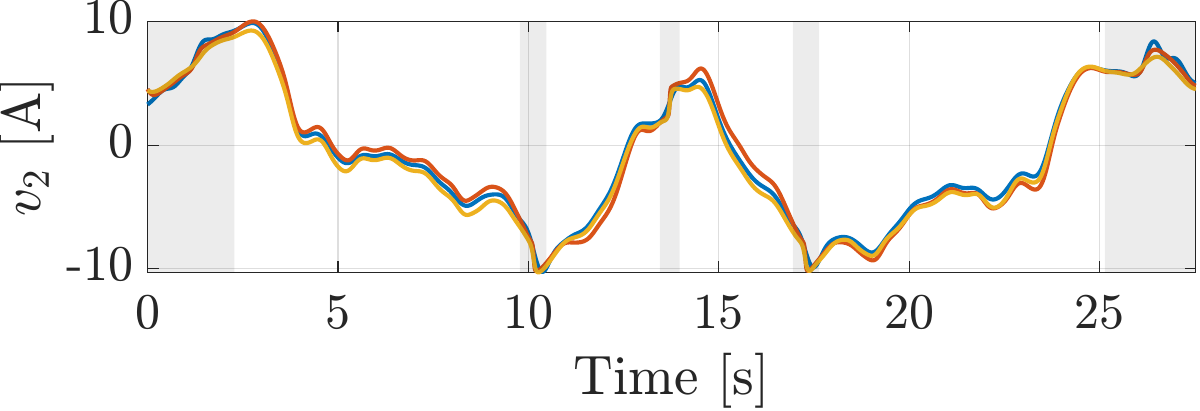}
    \end{subfigure}
    \hfill
    \begin{subfigure}[b]{\figsize}
        \centering
        \includegraphics[width=\textwidth]{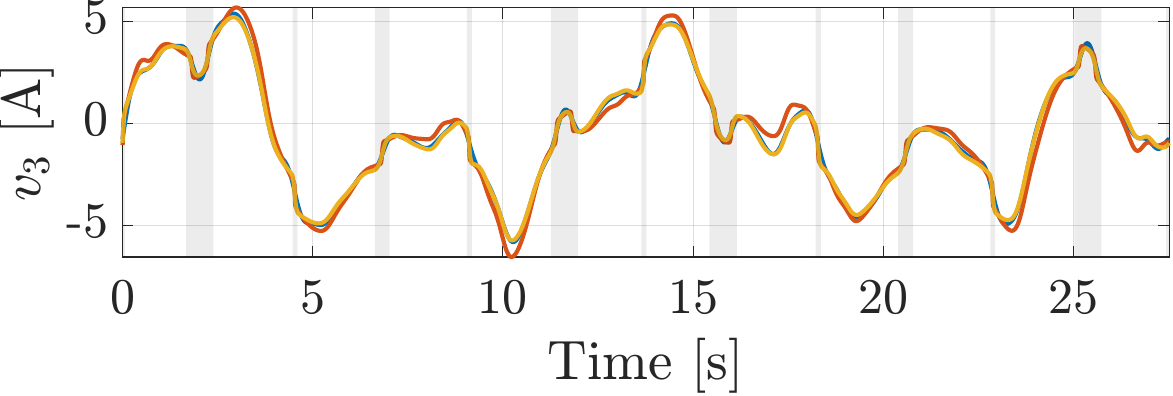}
    \end{subfigure}
    \\ \vspace{0.25em}
    \begin{subfigure}[b]{\figsize}
        \centering
        \includegraphics[width=\textwidth]{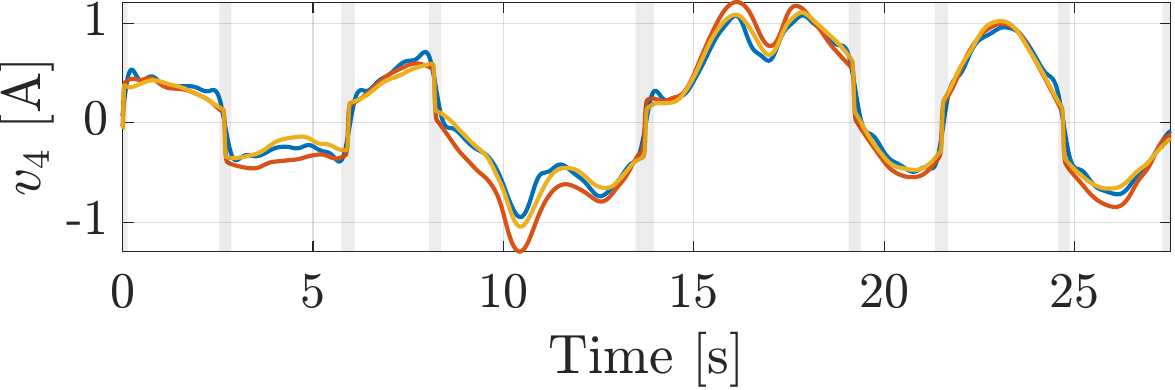}
    \end{subfigure}
    \hfill
    \begin{subfigure}[b]{\figsize}
        \centering
        \includegraphics[width=\textwidth]{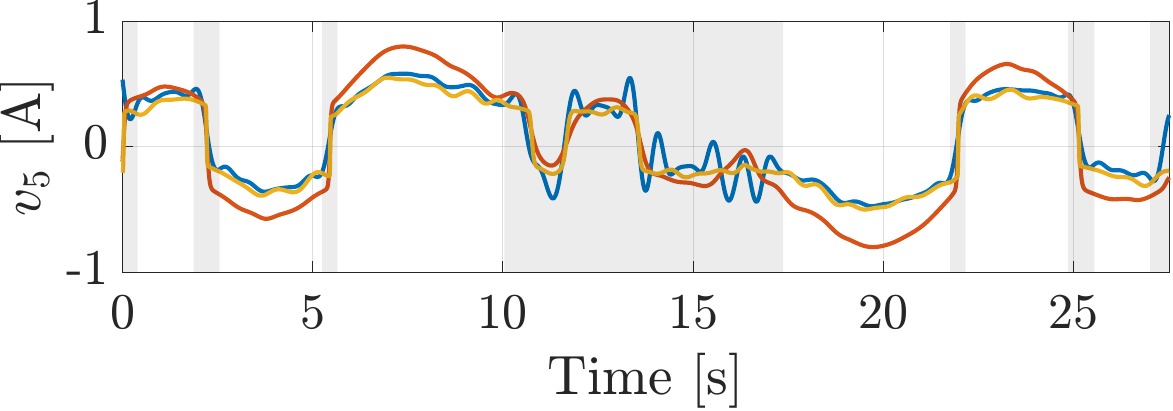}
    \end{subfigure}
    \hfill
    \begin{subfigure}[b]{\figsize}
        \centering
        \includegraphics[width=\textwidth]{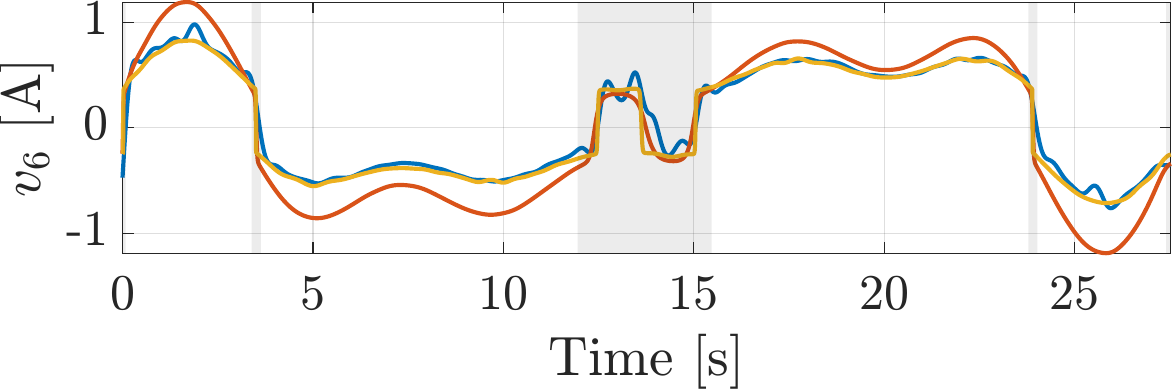}
    \end{subfigure}    

    \caption{Measured and estimated joint currents in Trajectory~A (first two rows) and Trajectory~B (last two rows)}
    \label{fig:comparison}
\end{figure*}

\begin{table}
\centering
\caption{$\MNAE(\bm v_j, \hat{\bm v}_j)$ in validation trajectories}
\label{tab:comparison}
\begin{tabular}{|c|ccc|ccc|} 
\hline
\multirow{2}{*}{\textbf{Joint}} & \multicolumn{3}{c|}{\textbf{Trajectory A}} & \multicolumn{3}{c|}{\textbf{Trajectory B}} \\
 & \cite{gaz_model-based_2018} & ours & $\eta$ & \cite{gaz_model-based_2018} & ours & $\eta$ \\ 
\hline
Joint 1 & 23.2838 & 10.0031 & 2.3277 & 15.6990 & 6.0429 & 2.5979 \\
Joint 2 & 5.4332 & 3.6834 & 1.4751 & 4.8266 & 4.4326 & 1.0889 \\
Joint 3 & 5.6983 & 2.1738 & 2.6214 & 5.0099 & 1.5387 & 3.2559 \\
Joint 4 & 10.3336 & 5.9348 & 1.7412 & 9.8104 & 5.6306 & 1.7423 \\
Joint 5 & 35.7345 & 8.0699 & 4.4281 & 32.7093 & 11.1431 & 2.9354 \\
Joint 6 & 28.3407 & 8.3120 & 3.4096 & 23.5660 & 5.4544 & 4.3205 \\
\hline
\end{tabular}
\end{table}

Fig.~\ref{fig:comparison} shows the comparison between the joint currents estimated by \cite{gaz_model-based_2018} and the currents estimated with our method, defined in Section~\ref{sec:linear-dynamic-parameters} and \ref{sec:nonlinear-friction-parameters}.
Table~\ref{tab:comparison} presents the numerical evaluation: as confirmed by the plots, we achieve a lower prediction error across all joints against both the validation trajectories, manifesting the superior performance of our model and demonstrating the efficacy of a dynamic estimation of $\bm v$ against a static retrieval of $\gls{gravity}(\bm q)$ only.
Remarkably, we achieve a maximum improvement factor of $\gls{improvement-factor} = 4.4281$ and $\gls{improvement-factor} = 4.3205$ in trajectories A and B respectively, with $\gls{improvement-factor} > 1$ for all the joints in both trajectories.

We point out that larger prediction errors are due to nonlinear effects: consider, for instance, the nonlinear regions in $t \in [7.5, 10]\si{\second}$ for Joint~6 in Trajectory~A, or in $t \in [12.5, 15]\si{\second}$ for Joint~6 in Trajectory~B.
Indeed, for $\lvert \dot q_6 \rvert < \overline{\dot q}$, we attained an $\MNAE$ index of $25.8714$ and $26.8361$, respectively, which is considerably higher than the same index computed for all samples.
Notably, this phenomenon is more evident for the last joints, for which the friction contribution plays a substantial role on the overall dynamics, compared to other effects.

\subsubsection{Nonlinear Friction Parameters}

\begin{figure}
    \centering
    \newcommand*{\figsize}{0.49\columnwidth}

    \begin{center}
    \footnotesize{{\legblue} measured \hspace{0.5em}{\legred} estimated \hspace{0.5em} {\boxgray} nonlinearity region}
    \end{center}

    \begin{subfigure}[b]{\figsize}
        \centering
        \includegraphics[width=\textwidth]{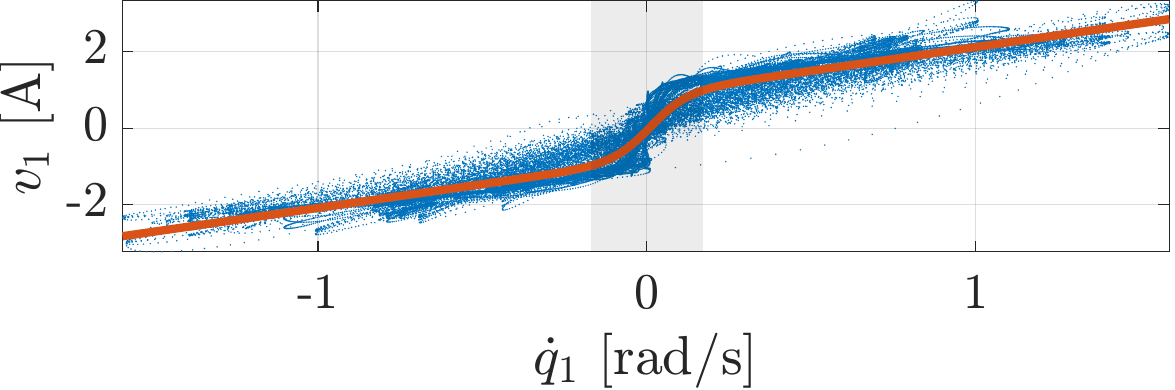}
    \end{subfigure}
    \hfill
    \begin{subfigure}[b]{\figsize}
        \centering
        \includegraphics[width=\textwidth]{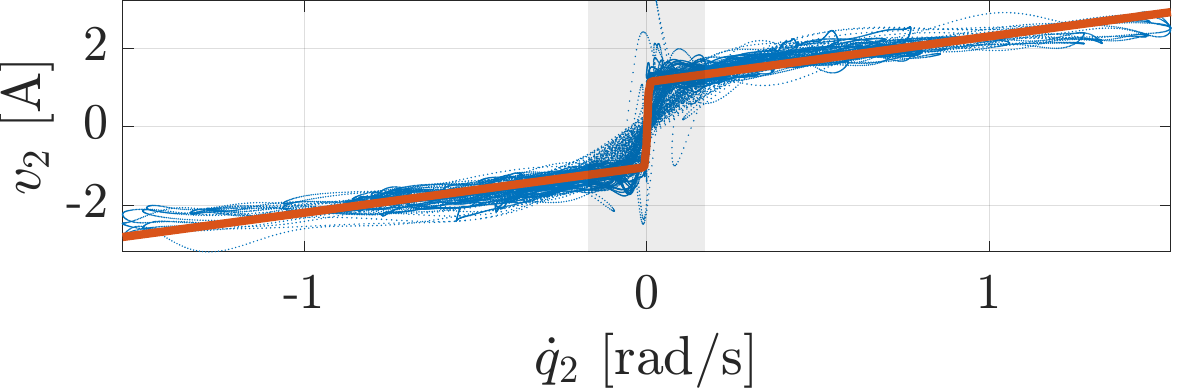}
    \end{subfigure}
    \\ \vspace{0.25em}
    \begin{subfigure}[b]{\figsize}
        \centering
        \includegraphics[width=\textwidth]{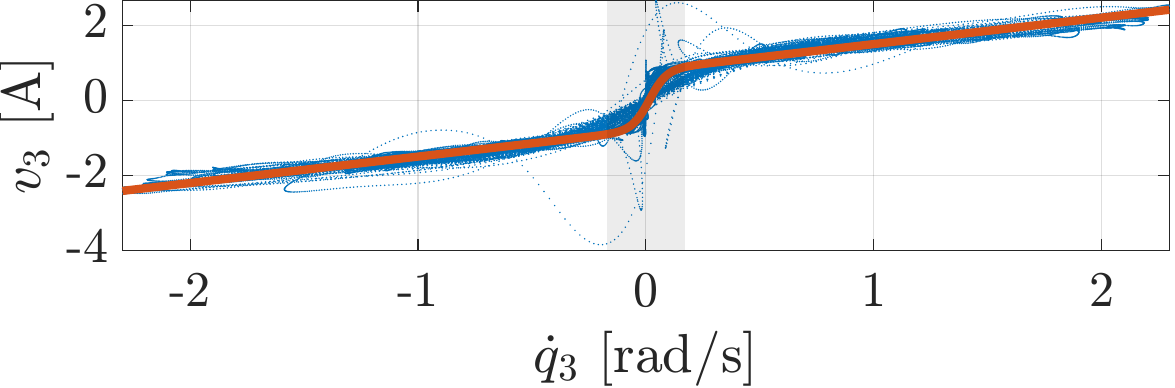}
    \end{subfigure}
    \hfill
    \begin{subfigure}[b]{\figsize}
        \centering
        \includegraphics[width=\textwidth]{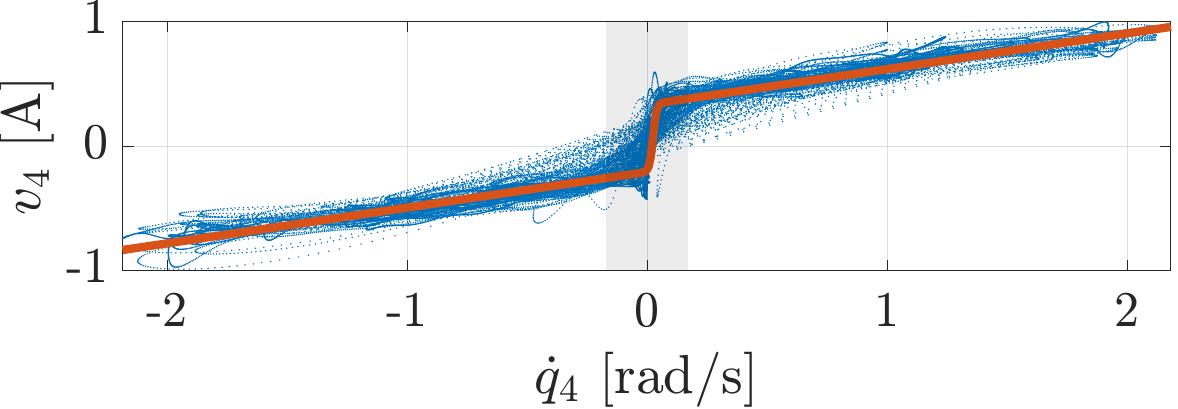}
    \end{subfigure}
    \\ \vspace{0.25em}
    \begin{subfigure}[b]{\figsize}
        \centering
        \includegraphics[width=\textwidth]{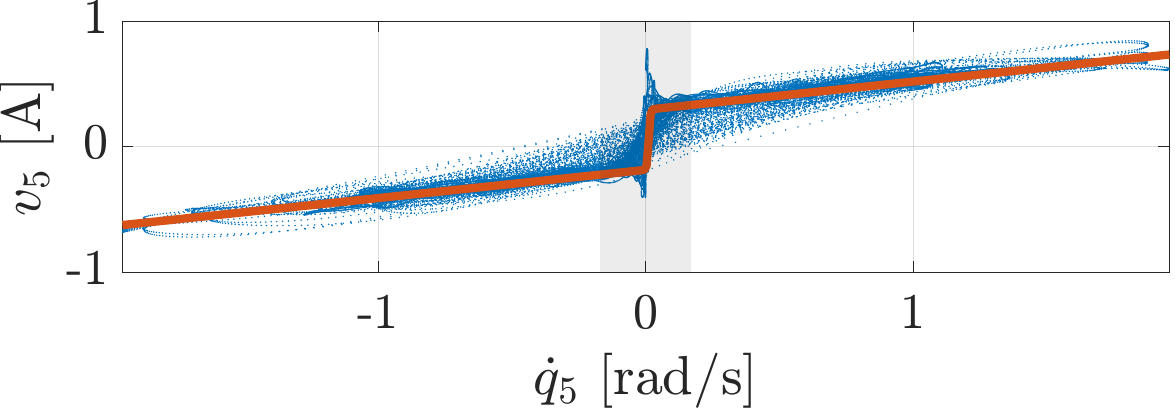}
    \end{subfigure}
    \hfill
    \begin{subfigure}[b]{\figsize}
        \centering
        \includegraphics[width=\textwidth]{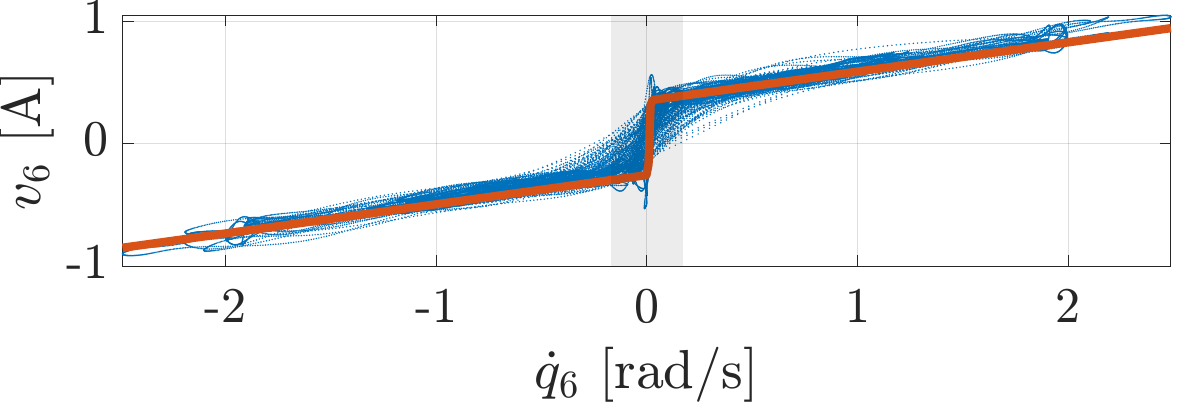}
    \end{subfigure}

    \caption{Measured and estimated joint friction currents}
    \label{fig:friction}
\end{figure}

\begin{table}
\centering
\caption{Estimated nonlinear friction parameters}
\label{tab:friction}
\begin{tabular}{|c|c|c|c|c|c|} 
\hline
\textbf{Joint} & $\bm f_v$ & $\bm f_o$ & $\bm f_c$ & $\bm\alpha$ & $\bm\nu$ \\ 
\hline
Joint 1 & 1.0640 & -1.0066 & 2.0506 & 7.9467 & -0.0185 \\
Joint 2 & 0.9944 & 0.9563 & -2.4017 & -59.9536 & -0.0019 \\
Joint 3 & 0.6796 & -0.8120 & 1.6478 & 19.8251 & -0.0053 \\
Joint 4 & 0.3159 & -0.1767 & 0.4688 & 134.8982 & -0.0186 \\
Joint 5 & 0.2244 & -0.1924 & 0.4760 & 331.4421 & -0.0118 \\
Joint 6 & 0.2358 & -0.2453 & 0.5980 & 459.1933 & -0.0130 \\
\hline
\end{tabular}
\end{table}

After estimating the linear parameters, the nonlinear friction parameters are estimated by using the procedure described in Section~\ref{sec:nonlinear-friction-parameters} and reported in Table~\ref{tab:friction}.
The friction currents are illustrated in Fig.~\ref{fig:friction}.
It is evident from the plots that $\underline{\hat{\bm v}}_f(\dot{\bm q})$ behaves nonlinearly for $\lvert \dot{\bm q} \rvert < \dot q^+$, hence a sigmoidal model \eqref{eq:friction-model} is fitted on them, while they tend to follow the linear model in \eqref{eq:linear-friction-model} for $\lvert \dot{\bm q} \rvert > \dot q^+$.

\subsubsection{Motor Drive Gains}

\begin{table}
\centering
\caption{Estimated joint motor drive gains [\si{\newton\meter\per\ampere}]}
\label{tab:motor-drive-gains}
\begin{tabular}{|c|c|c|c|c|} 
\hline
\textbf{Joint} & \textbf{GT} & \textbf{\cite{gaz_model-based_2018}} & \textbf{\cite{xu_robot_2022}} & \textbf{ours} \\ 
\hline
Joint 1 & 13.9557 & 14.87 & 14.7336 & 13.5841 \\
Joint 2 & 13.8669 & 13.26 & 14.3300 & 14.2959 \\
Joint 3 & 11.5049 & 11.13 & 11.5476 & 11.3716 \\
Joint 4 & 11.5438 & 10.62 & 11.2487 & 11.2408 \\
Joint 5 & 11.6143 & 11.03 & 11.5000 & 11.7682 \\
Joint 6 & 11.4149 & 11.47 & 11.5000 & 11.7681 \\ 
\hline
\textbf{MSE \wrt GT} & 0 & 1.5946 & 0.9637 & 0.7617 \\
\hline
\end{tabular}
\vspace*{-5mm}
\end{table}

To estimate the motor drive gains $\gls{mdg}$, we exercise the procedure described in Section~\ref{sec:motor-drive-gains}, originally developed in \cite{xu_robot_2022}, by attaching an eccentrically mounted payload, according to the EE frame shown in Fig.~\ref{fig:gripper}.
More specifically, $j^* = \gls{n-dof} - 1$, \ie, $\bm S_{\gls{n-dof}-1}$ and $\bm S_{\gls{n-dof}}$ are rank-deficient for the UR10 kinematic structure, hence \eqref{eq:estimated-motor-drive-gains} is solved for gains $K_{\gls{n-dof}-1}$ and $K_{\gls{n-dof}}$.

Table~\ref{tab:motor-drive-gains} reports the joint motor drive gains estimated in \cite{gaz_model-based_2018} and \cite{xu_robot_2022}, as well as our estimation.
On the one hand, \cite{gaz_model-based_2018} uses a procedure foreseeing a sequence of 500 static configurations as data samples to estimate the motor gains, based on the difference between the measured currents with and without holding an eccentric payload; this requires $K_1$ to be estimated separately.
On the other hand, we make use of a dynamic procedure, as in \cite{xu_robot_2022}, \ie we collect data from the execution of a set of trajectories, allowing a one-shot estimation of $\gls{mdg}$.

We compare our estimated gains with those obtained by the aforementioned baselines.
As ground-truth (GT), we measure a certain number $\gls{n-static-samples} \in \mathbb N$ of samples in different static configurations, and compute the motor gains as
\begin{equation}
    K_{GT,j} = \frac{1}{\gls{n-static-samples}} \sum_{k=1}^{\gls{n-static-samples}}{\frac{\tau_{d,j}^{(k)}}{v_{d,j}^{(k)}}}, \qquad \forall j \in \{ 1, \ldots, n \},
\end{equation}
where $\tau_{d,j}^{(k)}$ and $v_{d,j}^{(k)}$ indicate the $k$-th sample of, respectively, the target torques and currents commanded to the joints, both acquired through the software provided by the manufacturer.

The results collected in Table~\ref{tab:motor-drive-gains} show that, although a significant improvement is already achieved by \cite{xu_robot_2022} compared to \cite{gaz_model-based_2018}, an additional increase of accuracy is attained by replicating  \cite{xu_robot_2022} on our robot.
This suggests the necessity of performing the motor drive gains estimation from scratch for each industrial manipulator, as $\gls{mdg}$ might differ because of unavoidable manufacturing inaccuracies in the motors production.

\subsection{Payload Integration}\label{sec:experimental-results-payload-integration}

\begin{figure*}
    \centering
    \newcommand*{\figsize}{0.32\textwidth}

    \begin{center}
    \footnotesize{{\legblue} computed \hspace{0.5em} {\legred} \ref{itm:scenario-2} \hspace{0.5em} {\legyellow} \ref{itm:scenario-3}}
    \end{center}

    \begin{subfigure}[b]{\figsize}
        \centering
        \includegraphics[width=\textwidth]{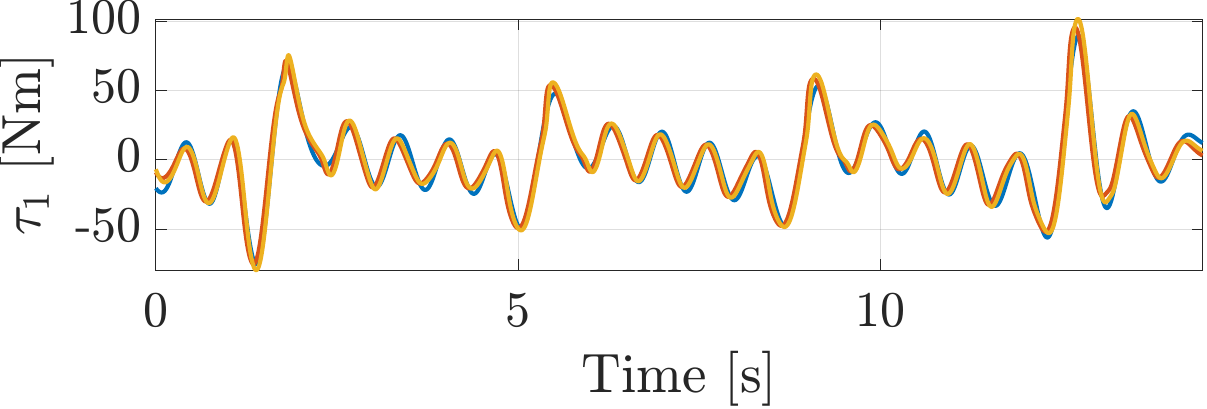}
    \end{subfigure}
    \hfill
    \begin{subfigure}[b]{\figsize}
        \centering
        \includegraphics[width=\textwidth]{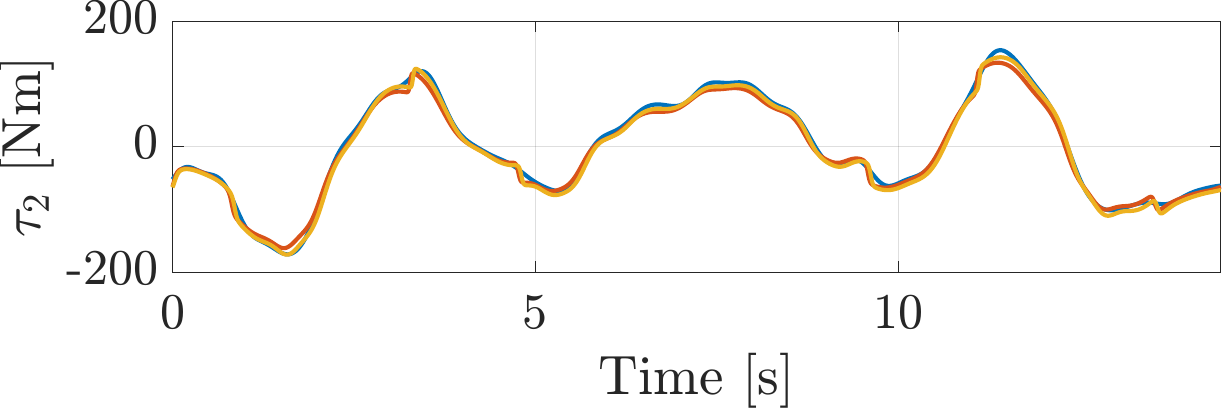}
    \end{subfigure}
    \hfill
    \begin{subfigure}[b]{\figsize}
        \centering
        \includegraphics[width=\textwidth]{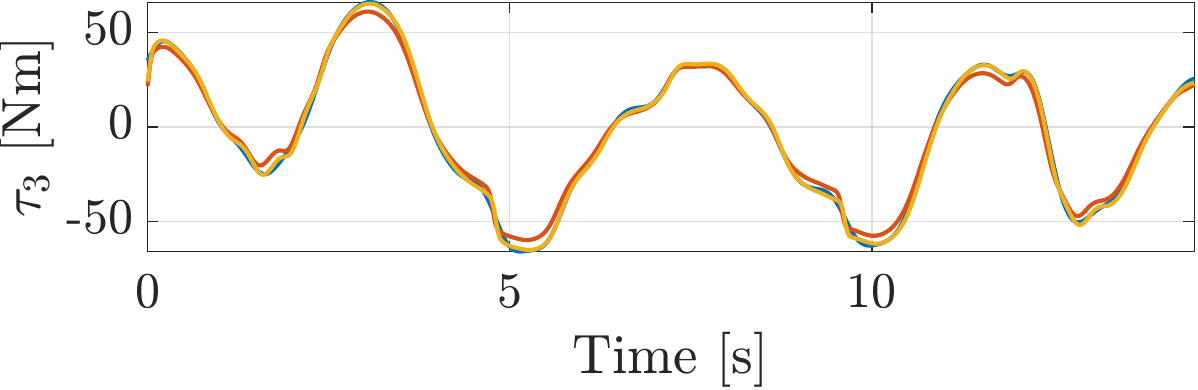}
    \end{subfigure}
    \\ \vspace{0.25em}
    \begin{subfigure}[b]{\figsize}
        \centering
        \includegraphics[width=\textwidth]{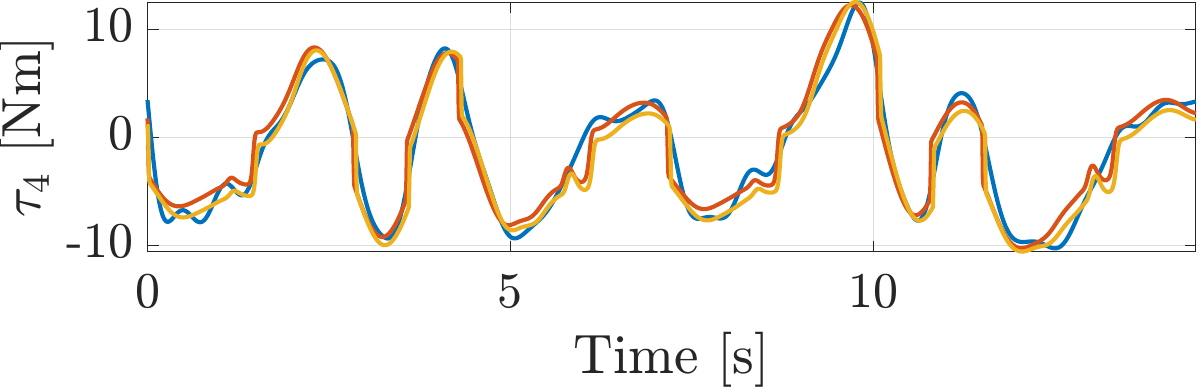}
    \end{subfigure}
    \hfill
    \begin{subfigure}[b]{\figsize}
        \centering
        \includegraphics[width=\textwidth]{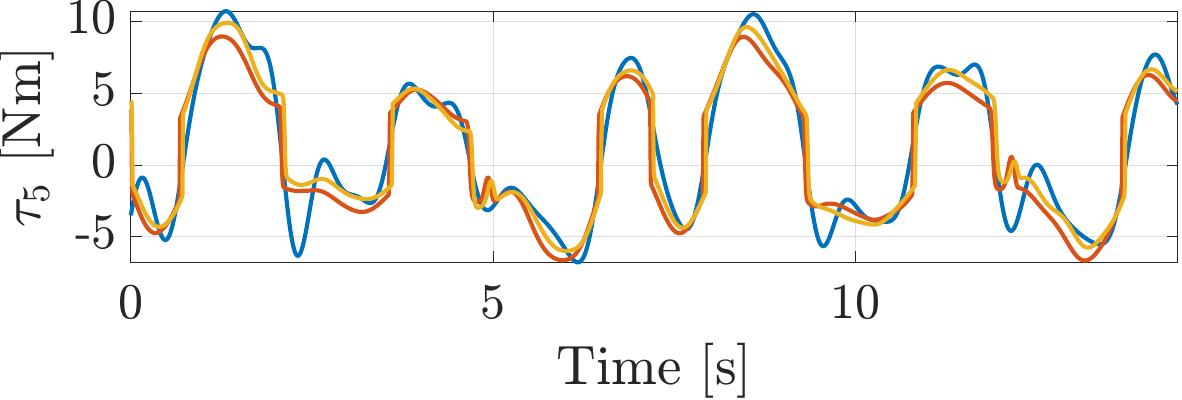}
    \end{subfigure}
    \hfill
    \begin{subfigure}[b]{\figsize}
        \centering
        \includegraphics[width=\textwidth]{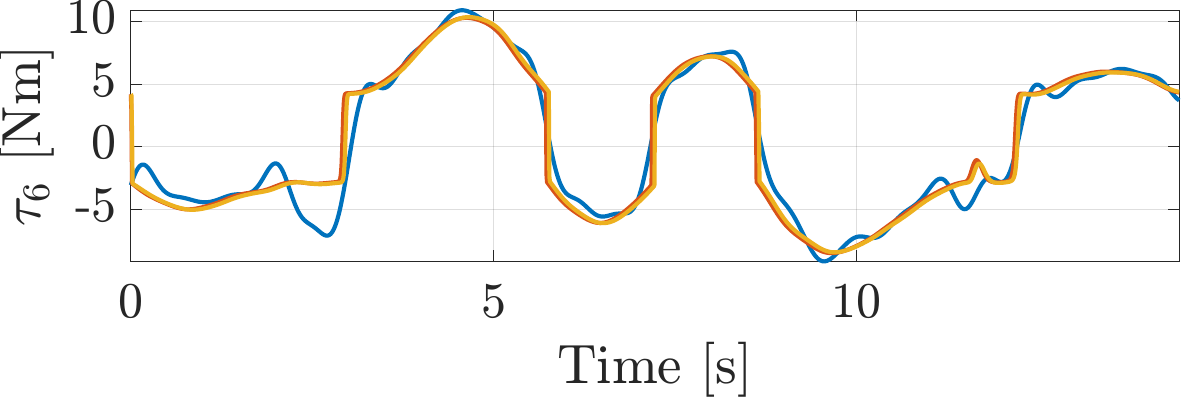}
    \end{subfigure}

    \caption{Estimated joint torques compared to torques computed from measured currents}
    \label{fig:payload}
\end{figure*}

\begin{table}
\centering
\caption{$\MNAE(\bm \tau_j, \hat{\bm \tau}_j)$ obtained by validating the dynamics solver against different payload configurations}
\label{tab:payload}
\begin{tabular}{|c|c|c|c|} 
\hline
\textbf{Joint} & \textbf{\ref{itm:scenario-1}} & \textbf{\ref{itm:scenario-2}} & \textbf{\ref{itm:scenario-3}} \\ 
\hline
Joint 1 & 4.3672 & 4.0876 & 4.1121 \\
Joint 2 & 3.3818 & 3.8360 & 3.1968 \\
Joint 3 & 3.0961 & 4.6370 & 1.5898 \\
Joint 4 & 9.1918 & 8.3397 & 8.5546 \\
Joint 5 & 10.1428 & 13.7379 & 11.2278 \\
Joint 6 & 8.8053 & 8.2813 & 8.2736 \\ 
\hline
\textbf{Average} & 6.4975 & 7.1533 & 6.1591 \\
\hline
\end{tabular}
\end{table}

The integration of a payload attached to the EE flange is assessed with 3 experimental scenarios:
\begin{enumerate}[label={Scenario~\arabic*)}, ref={Scenario~\arabic*}, wide=\parindent]
    \item a trajectory is run without the robot holding a payload, the torques computed from measured currents $\bm\tau = \hat{\bm K} \bm v$ are compared with the estimations of the IDS configured without a payload;\label{itm:scenario-1}
    \item the trajectory is run with the robot holding a payload (\ie, the hand in Fig.~\ref{fig:franka-hand}), the torques computed from measured currents are compared with the estimations of the IDS configured without a payload;\label{itm:scenario-2}
    \item the trajectory is run with the robot holding the payload, the torques computed from measured currents are compared with the estimations of the IDS configured with the payload inertial parameters in Table~\ref{tab:franka-hand}.\label{itm:scenario-3}
\end{enumerate}

Equation \eqref{eq:tau-arm-tau-l} is implemented in our IDS with
\begin{equation}
    \bm R_n^l \coloneqq \begin{bmatrix}
        \nicefrac{\sqrt 2}{2} & -\nicefrac{\sqrt 2}{2} & 0 \\
        \nicefrac{\sqrt 2}{2} & \nicefrac{\sqrt 2}{2} & 0 \\
        0 & 0 & 1
    \end{bmatrix}, \qquad
    \bm t_n^l \coloneqq \begin{bmatrix}
        0 \\
        0 \\
        0
    \end{bmatrix},
\end{equation}
\ie Franka Hand is concentrically mounted with a \SI{45}{\degree} rotation around $z_6$, attached without any translational offset.

Table~\ref{tab:payload} shows that, compared to \ref{itm:scenario-1}, the estimation error increases in \ref{itm:scenario-2}, demonstrating the decrease of performance when the IDS ignores the actual presence of the payload.
In contrast, when the solver is properly configured in \ref{itm:scenario-3}, an improvement is attained.
Fig.~\ref{fig:payload} shows the measured and estimated joint torques in scenarios 2 and 3.
It is worth pointing out that differences in the performance between the scenarios might be more evident with a more dynamically exciting payload.
In fact, Franka Hand's mass ($< \SI{1}{\kilo\gram}$) is rather limited compared to the manipulator capacity ($\SI{10}{\kilo\gram}$).

\subsection{Software Implementation}\label{sec:software-implementation}

Our UR10 dynamics solver is a C++ software  library developed according to the ROS2 framework.
Once the modeling and identification procedure described in Section~\ref{sec:dynamic-model-identification} is performed on the UR10 robot, the estimated model implementing equations \eqref{eq:estimated-currents-with-friction}--\eqref{eq:estimated-torques-and-gains} is converted from MATLAB symbolic equations to C++ functions using the \texttt{matlabFunction} method and the MATLAB Coder Toolbox.

\subsubsection{Functionalities}\label{sec:functionalities}

The IDS library exposes functions to allow the user to extract $\gls{inertia}(\bm q)$, $\gls{coriolis}(\bm q,\dot{\bm q})\dot{\bm q}$, $\gls{friction}(\dot{\bm q})$, $\gls{gravity}(\bm q)$ or $\bm\tau$.
In particular, while $\bm\tau$ is computed according to \eqref{eq:estimated-torques-and-gains}, with $\bm v$ calculated as \eqref{eq:estimated-currents-with-friction}, all the dynamics terms are retrieved using ``partial'' regressors and dynamic coefficients, constructed considering only the portions of the EOM in \eqref{eq:eom} describing each particular term.
Mathematical details on how to implement the EOM can be found in \cite{siciliano_dynamics_2009}.
Moreover, it is possible to configure the solver with the payload parameters by reading a YAML file, a common format used by the ROS2 community.
A \textit{launch file} is responsible for both loading the configuration file and instantiating and configuring the IDS, which reads the payload parameters at instantiation time.  

\subsubsection{Architecture}

The software architecture is composed by an abstract class, named \texttt{InverseDynamicsSolver}, providing the functionalities mentioned in Section~\ref{sec:functionalities}: this class represents a generic solver, and the particular implementation of the methods returning dynamic components is a responsibility of the concrete classes.
The actual UR10's IDS, named \texttt{InverseDynamicsSolverUR10}, is one of the two concrete implementations of the abstract class we provide: this class returns the torques for the specific manipulator under study, according to the estimated model.
Additionally, the abstract class is inherited by \texttt{InverseDynamicsSolverKDL}, which computes the dynamic components through the KDL library, after reading the robot description from URDF files: this class is useful when a complete kinematic and dynamic description is available, \eg in the case both the DH and dynamic parameters are provided by the manufacturer.
Lastly, it is worth remarking that users in the ROS2 community are allowed to inherit the base class with their own implementations, possibly including the estimated models of other manipulators.

\subsubsection{Implementation details}

The IDS defines the input/output types of the implemented methods using \texttt{Eigen}, a popular C++ linear algebra  library.
This way, users can easily manipulate the computed matrices for model based applications, such as writing control schemes or evaluating planning algorithms \cite{petrone_time-optimal_2022}.
For further details, interested readers are referred to the supplementary code capsule.

\section{Conclusions}\label{sec:conclusions}

This paper introduced the dynamic model of the UR10 industrial robot, and presented the identification procedure performed to estimate the manipulator's dynamic coefficients.
The complete identification process is composed of three subsequent stages, to face several issues due to the structural nature of the robot.

The first step consists in devising linear dynamic coefficients to retrieve the estimated joint currents, according to a well-known linear least square solution, hence obtaining a linear current-level dynamic model, as joint torques are not directly measurable.
Since the joint currents exhibit a nonlinear behavior due to joint friction, the second phase requires establishing a nonlinear optimization problem, to fit a set of friction parameters to a sigmoidal model.
The third and last stage finally outputs joint torques: indeed, it concerns the estimation of motor drive gains, to convert the current-level dynamic model to the joint torque level.

Compared to the current-level model identified in \cite{gaz_model-based_2018}, we consistently achieved a lower estimation error on all the joints, validated against 2 exciting trajectories; however, the adopted procedure still presents the major flaw of lacking a global solution for the nonlinear optimization problem to devise friction parameters.
The estimated torque-level dynamic model has been implemented in a ROS2 software module we developed and released, that can be used off-the-shelf in a variety of control, planning, and simulation applications.
The IDS is validated in various scenarios, also foreseeing the inclusion of a payload, demonstrating that the software can be reconfigured to account for a generic known payload.

In future research, we aim to refine excitation trajectory design. Currently, reference joint sinusoid parameters are chosen heuristically to explore the robot's workspace.
In this sense, investigating the impact of optimal excitation trajectory computation \cite{luo_optimal_2023} on parameter estimation accuracy during validation would be a valuable improvement.
Alternatively, adopting online identification approaches could prove particularly beneficial by eliminating the need for designing trajectories tailored to meet the PE condition \cite{guo_composite_2019}, thereby simplifying the process and potentially enhancing efficiency.

Additionally, we intend to explore alternative friction models, such as those incorporating the Stribeck effect \cite{tadese_passivity_2021} or degressive friction \cite{weigand_dataset_2023}.
To address the challenge of globally optimizing nonlinear friction parameter estimation, we might employ resolution techniques like evolutionary algorithms \cite{ding_nonlinear_2018} or neural networks \cite{guo_composite_2019}.

Finally, training the model on data averaged across multiple identical trajectories, rather than relying on single trajectories with low-pass filtering, could enhance current prediction accuracy by mitigating measurement noise without distorting high-frequency motion signals.

\section*{Acknowledgment}
The authors would like to thank Giovanni Mignone and Antonio Annunziata for refining the implementation of the identification procedure, making this work possible.

\printunsrtglossary[type=symbols]\label{sec:list-of-symbols}

\bibliographystyle{IEEEtran}
\bibliography{IEEEabrv,OtherAbbrv,tii-2024-dynamic-identification}

\vspace*{\fill}
\begin{IEEEbiography}[{\includegraphics[width=1in,height=1.25in,clip,keepaspectratio]{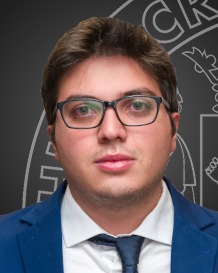}}]{Vincenzo Petrone}
(Student Member, IEEE) received the \ms degree in computer engineering with honors from the University of Salerno, Fisciano, Italy, in 2021.
He is currently working toward the \phd degree in information engineering with the University of Salerno, Fisciano,
Italy.
Between 2023 and 2024, he was a Visiting \phd Student with the Dalle Molle Institute for Artificial Intelligence (IDSIA), Lugano, Switzerland, University of Applied Sciences and Arts of Southern Switzerland (SUPSI), Manno, Switzerland, Università della Svizzera italiana (USI), Manno, Switzerland, for six months, focusing on reinforcement learning techniques applied to robotic interaction tasks.
His research interests cover optimal trajectory planning and interaction control of robotic tasks, parametric identification of dynamics models, and intelligent control.
\end{IEEEbiography}

\begin{IEEEbiography}[{\includegraphics[width=1in,height=1.25in,clip,keepaspectratio]{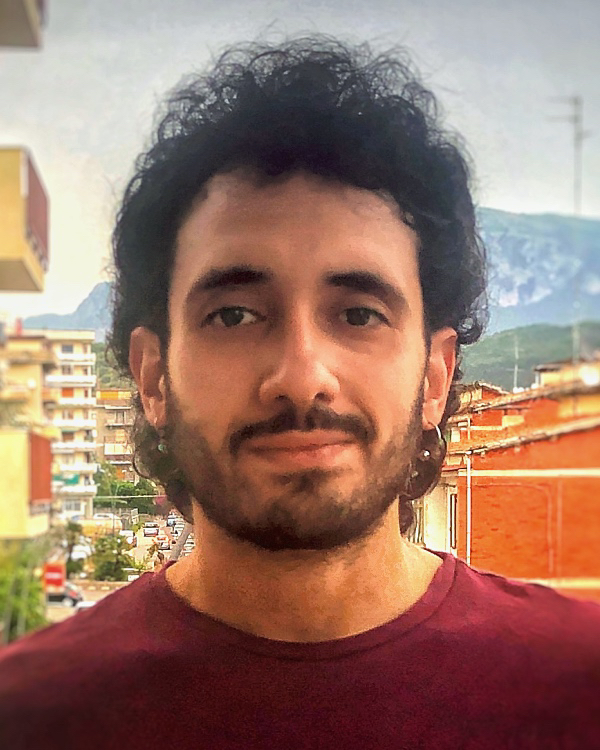}}]{Enrico Ferrentino}
(Member, IEEE) received the \ms degree in computer and automation
engineering from the Polytechnic University of Turin, Turin, Italy, in 2014, and the \phd degree in robotics from the University of Salerno, Fisciano, Italy, in 2020.
From 2013 to 2014, he was a Visiting Scholar with NASA’s Jet Propulsion Laboratory, Pasadena, CA, USA, where he contributed to the Axel tethered robot project within the Robotic Mobility group, led by \dr Issa Nesnas.
Between 2014 and 2017, he was a Ground Segment Engineer with ALTEC S.p.A., Turin, Italy, working on the ESA robotic mission ExoMars.
In 2019, he was a Visiting Scholar with LAAS-CNRS, Toulouse, France, where he participated in the H2020 PRO-ACT project as part of the team led by \prof Antonio Franchi.
Since 2020, he has been a Lecturer in Robotics and Medical Robotics with the Department of Information Engineering, Electrical Engineering and Applied Mathematics, University of Salerno, where he is currently an Assistant Professor and technical advisor of the Robotics Laboratory, led by \prof Pasquale Chiacchio.
In 2024, he was appointed Associate Editor for the IEEE Transactions on Automation Science and Engineering.
His research interests include optimal planning and control of robotic manipulators, force control, haptic teleoperations, human–robot collaboration, and software architectures for robot control, with applications spanning industrial, medical, and aerospace robotics.
\end{IEEEbiography}

\begin{IEEEbiography}[{\includegraphics[width=1in,height=1.25in,clip,keepaspectratio]{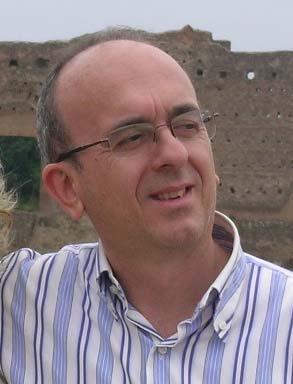}}]{Pasquale Chiacchio}
(Senior Member, IEEE) received the \ms degree in electronic engineering and the \phd degree in electronic and information engineering from University of Naples Federico II, Naples, Italy, in 1987 and 1992, respectively.
He is currently a Professor in Automatic Control and Robotics with the Department of Information Engineering, Electrical Engineering and Applied Mathematics, University of Salerno, Fisciano, Italy.
His main research interests include robotics and modeling and control of discrete event systems.
In the robotics field, he has been working on robot control and identification, inverse kinematics problem, interaction control, control of redundant manipulators, and control of cooperative manipulators.
In the discrete event systems field, he has been working on supervisory control based on monitors, optimal supervisory control, and formal specification for supervisory systems.
The results have been published in the main journals of the sector and have been accompanied by an intense experimental activity.
He has been a Coordinator of two Research Projects of National Interest (PRIN) and has been involved in research projects funded by the European Union (ECHORD, AIRobots, EuRoC, LOCOMACHS, and LABOR).
In December 2011, he was nominated Knight and then in July 2014, he was promoted to an Officer of the Order of Merit of the Italian Republic.
Since 2016, he has been the Director of the \phd Program in Information Engineering.
\prof Chiacchio is a member of the Italian Society of Control
Researchers.
\end{IEEEbiography}
\vspace*{\fill}

\end{document}